\documentclass[lettersize,journal]{IEEEtran}
\usepackage{amsmath,amsfonts}
\usepackage{algorithmic}
\usepackage{algorithm}
\usepackage{array}
\usepackage[caption=false,font=normalsize,labelfont=sf,textfont=sf]{subfig}
\usepackage{textcomp}
\usepackage{url}
\usepackage{verbatim}
\usepackage{graphicx}
\usepackage{cite}

\usepackage{amsfonts}
\usepackage{xcolor}

\definecolor{red_old}{RGB}{237, 103, 90}
\definecolor{green_new}{RGB}{51, 209, 122}
\definecolor{blue_retro}{RGB}{26, 95, 180}

\definecolor{gamboge}{RGB}{243, 156, 18}
\definecolor{celeste}{RGB}{52, 152, 219}
\definecolor{charcoal}{RGB}{44, 62, 80}

\usepackage{ bbold } 

\usepackage[normalem]{ulem} 

\usepackage{multirow} 

\usepackage{makecell} 

\usepackage{orcidlink}

\begin{document}

\title{Continual Low-Rank Scaled Dot-product Attention}

\author{Gin\'es Carreto Pic\'on, Illia Oleksiienko, Lukas Hedegaard, Arian Bakhtiarnia, and Alexandros Iosifidis
\thanks{Email addresses: gcp@ece.au.dk (G. Carreto Picón), io@ece.au.dk (I. Oleksiienko), lhm@ece.au.dk (L. Hedegaard), arianbakh@ece.au.dk (A. Bakhtiarnia), alexandros.iosifidis@tuni.fi (A. Iosifidis)}
\thanks{G. Carreto Pic\'on, I. Oleksiienko, L. Hedegaard, A. Bakhtiarnia and A. Iosifidis are with the Department of Electrical and Computer Engineering, Aarhus University, Denmark.}
\thanks{A. Iosifidis is with the Faculty of Information Technology and Communciation Sciences, Tampere University, Finland}
\thanks{G. Carreto Pic\'on and A. Iosifidis acknowledge funding by the Horizon Europe programme PANDORA (GA 101135775)}
\thanks{The code produced for this work is available in \url{https://github.com/Atamarado/continual_nystrom}}
}

\maketitle

\begin{abstract}
Transformers are widely used for their ability to capture data relations in sequence processing, with great success for a wide range of static tasks. However, the computational and memory footprint of their main component, i.e., the Scaled Dot-product Attention, is commonly overlooked. This makes their adoption in applications involving stream data processing with constraints in response latency, computational and memory resources infeasible. Some works have proposed methods to lower the computational cost of Transformers, i.e. low-rank approximations, sparsity in attention, and efficient formulations for Continual Inference. In this paper, we introduce a new formulation of the Scaled Dot-product Attention based on the Nyström approximation that is suitable for Continual Inference. In experiments on Online Audio Classification and Online Action Detection tasks, the proposed Continual Scaled Dot-product Attention can lower the number of operations by up to three orders of magnitude compared to the original Transformers while retaining the predictive performance of competing models.
\end{abstract}


\section{Introduction}
\label{sec1}

\noindent\IEEEPARstart{T}{ransformers} \cite{Vaswani17transformer} are a general purpose model with a wide range of applications, including Machine Translation \cite{Vaswani17transformer}, Natural Language Processing (NLP) \cite{Patwardhan23nlp}, Genome Sequencing \cite{Choi23genome} and Computer Vision \cite{Ulhaq22ar_survey}. The core component of the Transformer is the Scaled Dot-product Attention (SDA), which receives three matrices $Q$, $K$ and $V$ as inputs, where each row of which corresponds to a token introduced to the SDA, computed by the linear transformations $Q = X_Q W_Q$, $K = X_K W_K$ and $V = X_V W_V$. In the case of self-attention, all three input matrices are identical, i.e., $X_Q=X_K=X_V$, while when SDA is used to implement cross-attention of two inputs, $X_K=X_V$ corresponds to the first and $X_Q$ to the second input data to be fused by the SDA. $Q$, $K$ and $V$ are then used to perform the following transformation:

\begin{equation}
    \text{Att}(Q, K, V) = \text{s}\left(\frac{QK^T}{\sqrt{d}}\right) V,\label{eq:SDA}
\end{equation}

where $\text{s}(\cdot)$ is the softmax function applied row-wise on its input. The above computations on the input matrices $X_Q$, $X_K$ and $X_V$ are commonly performed multiple times in parallel, leading to multiple SDA-based data transformations, often referred to Multi-Head Attention. Thanks to this attention mechanism, the model can establish relations between the different tokens in the input sequence, which is the key aspect giving Transformers a great capacity to solve a wide variety of tasks. Even though their original formulation targeted language data, it has been shown that other types of data like images can be processed by Transformer models through tokenization, as done in the Vision Transformers \cite{Dosovitskiy21ViT}.

Each of the three input matrices in Eq. (\ref{eq:SDA}) has a dimensionality of $n \times d$, where $n$ is the number of tokens and $d$ the number of features, leading the computational and memory costs of the SDA module to have a quadratic form $O(n^2d)$. For large $n$, this attention mechanism can be too expensive, both computationally and memory-wise, for certain tasks where model inference needs to be performed with limited resources, such as tasks coming from applications in Robotic Perception and Control \cite{Aggarwal22robotics, Brohan23robotics}, Forecasting \cite{Miller24forecasting}, and Computer Vision \cite{bakhtiarnia2022single, bakhtiarnia2024efficient}.

Different approaches have been proposed to improve the efficiency of Transformer-based models. The simplest way is to reduce the size of the model and, in particular, the number of Transformer layers and attention heads \cite{Fan20transformer_depth}. However, due to the much lower number of learnable parameters, the resulting model may have insufficient capacity to perform well on complex tasks. This creates the need for structural model modifications that reach a better compromise between learning capacity and resource requirements. One approach for reducing the overall cost of the attention product is to reduce the number of features per token \cite{Xu24PCAFormer, Zhou21Informer}. Another approach is to limit the attention window of each token, as done by the Sparse Transformer \cite{Child19sparse_transformers} which reduces the attention window of each token to $\sqrt{n}$ adjacent tokens, leading to a computational cost of $O(n\sqrt{n})$. The Longformer \cite{Beltagy20longformer} generalizes this concept by using a small fixed window around each token to capture local patterns and a dilated sliding window (where not all tokens are used) to perform the attention efficiently.
The Performer \cite{Choromanski21performer} approximates the softmax attention with a Gaussian kernel by selecting a set of orthogonal random features to represent the entirety of the attention window. A different approach is to use low-rank matrix approximation schemes \cite{Drineas05nystrom, Yu17low_rank} for reducing both the number of computations and the size of the matrices involved in the SDA, as done by the Nyströmformer \cite{Xiong21nystromformer} which approximates the matrix multiplication followed by softmax nonlinearity in SDA. 
Efficient designs of Transformer model architectures also include the aggregation of neighboring tokens \cite{Yuan21ViT_training}, the addition of convolutional layers to reduce the input dimensionality \cite{Wu21CvT}, and architectures which tackle challenges of processing videos \cite{Khan22vision_transformers} such as using different restricted attentions \cite{Bertasius21space_time_attention, Liu22video_swin}, or aggregating features to reduce the number of tokens in the sequence \cite{Arnab21ViViT, Fan21multiscaleViT}.

While efficient model architectures have been proposed that can lead to fast inference for static tasks, such as the classification of images, videos or audio sequences, Continual Inference tasks remain challenging for Deep Learning models (even for those models that can operate in real time on the corresponding static task). Continual Inference \cite{Hedegaard22continual_inference} can be defined as the process of providing a result for each input of a continual data stream, for instance performing event/action classification based on the visual frames captured by a camera operating continually for a long period of time. An inherent requirement for solutions targeting Continual Inference tasks is that they need to be able to process the incoming stream of data with low latency and low resource consumption. 

Some model components are inherently continual by their design, as they accumulate their previous results into a memory. Some examples of this are Recurrent Neural Networks \cite{Schmidt19RNNs, Shertinsky18LSTM} and, more recently, Mambas \cite{Gu23Mambas}. However, neither of these models has yet replicated the success of Transformers \cite{Vaswani17transformer} in such a great variety of tasks.

The adoption of Transformer models, including the original and approximate formulations as well as the efficient architectures mentioned above, in a Continual Inference setting requires a transformation of the stream data processing task to its corresponding static task every time inference is performed. This is usually achieved with a sliding temporal window that creates a sequence segment (e.g., when processing visual streams, a video clip) formed by the newly captured data frame and the data frames preceding it. The full sequence within the window is then passed through the model to provide an inference result. This process leads to a very high amount of redundant computations when performing inference over successive windows, as every input data frame needs to be processed more than once (specifically, $n$ times for a sliding window with size $n$ data frames and step size of one). Continual Transformers \cite{Hedegaard23cont_trans} were proposed for addressing this limitation by processing each input token once. This is done by caching and reusing intermediate results to eliminate redundant computations in sequence data processing, following the principles of Continual Inference Networks (CINs) \cite{Hedegaard22continual_inference, Hedegaard23cont_trans, Hedegaard23GCNN, Hedegaard22Co3DCNN}.

Continual Inference methods are essentially redundant-free versions of their corresponding non-continual counterparts that produce identical outputs \cite{Hedegaard23cont_trans}. At the same time, the Nyström-based SDA formulation \cite{Xiong21nystromformer} has been extensively used in previous works \cite{Jeevan21xformers} as an alternative to the Transformer with lower computational cost. The adaptation of this SDA formulation to a Continual Inference setting requires a comprehensive mathematical analysis. This can be used as a baseline for the continual adaptation of other low-rank approximation schemes, as the challenges are expected to be similar. In this paper, we make the following contributions:
\begin{itemize}
    \item We propose a new formulation of the Continual Transformer, which further improves its memory and computational cost requirements for Continual Inference settings. We incorporate the Nyström-based approximation of the matrix multiplication followed by the softmax nonlinearity in SDA, to further lower the Continual SDA's memory footprint and the number of computations compared to its original formulation. To do this, we derive the model updates of the Nyström-based SDA in a continual manner.
    \item We propose two different ways to determine the landmarks used for processing continual stream data in the SDA approximation, and make the corresponding module modifications.
\end{itemize}

The remainder of the paper is structured as follows. Section \ref{sc:no} presents some helpful notation for describing models tailored to Continual Inference setting and used thereafter for describing the proposed model and its updates. Section \ref{sc:rw} provides an overview of related prior work. Section \ref{sc:me} describes the proposed Continual Transformer model. Experimental results and performance comparisons on Audio Classification and Online Action Detection tasks are provided in Section \ref{sc:ex}, and the conclusions are drawn in Section \ref{sc:co}.

\section{Notations useful for stream data processing} \label{sc:no}
Let us assume that a matrix $\Omega \in \mathbb{R}^{n \times d}$ formed by $n$ sequence tokens is updated in a continual manner, i.e., its top-most row corresponds to the oldest sequence token and its bottom-most row corresponds to the newest sequence token. When an update takes place, we define two tokens, i.e., $\omega_\text{old}$, which corresponds to the oldest token included in $\Omega$ before the update, and $\omega_\text{new}$, which is the new token to be included by the update. Then:
\begin{itemize}
    \item $\Omega_\text{mem}$ is formed by the $n-1$ tokens already included in $\Omega =\begin{bmatrix} \omega_\text{old} \\ \Omega_\text{mem} \end{bmatrix}$ which shift positions in the sequence such that after the update we have $\Omega = \begin{bmatrix} \Omega_\text{mem} \\ \omega_\text{new} \end{bmatrix}$.
    
    \item When an update takes place and $\Omega_\text{mem}$ needs to be updated to incorporate the influence of $\omega_\text{new}$, we define $\hat{\Omega}$ as $\Omega_\text{mem}$ before the update.
 \end{itemize}

The notation above can also be used for vectors (represented with lowercase letters) where the corresponding \textit{new} and \textit{old} elements will correspond to single values.

\section{Background work} \label{sc:rw}
The proposed method targets (approximate) Transformer inference with lower computational and memory requirements. This is done by adapting the Nyström-based approximation of the matrix multiplication followed by softmax nonlinearity so that it can be performed in a Continual Inference setting \cite{Hedegaard22continual_inference}. In the following, we provide an overview of the Continual Transformer and the Nyström-based formulation of SDA, which form the basis of our work. 

\subsection{Continual Transformer Attention}
The Continual Transformer \cite{Hedegaard23cont_trans} adapts the formulation of the SDA in Eq. (\ref{eq:SDA}) for Continual Inference settings, leading to the Continual Retroactive Attention formulation, which reuses computations performed at prior inference steps. The Continual Retroactive Attention is defined as follows:
\begin{equation} \label{eq:exp_trans}
    \text{Att}_\text{CoRe}(q_\text{new},k_\text{new},v_\text{new})=\phi(A)^{-1} \odot AV,
\end{equation}
where: 
\begin{align}
    \phi(A) &= A\mathbb{1}_n^T \in \mathbb{R} ^ {n \times 1}, \label{eq:phi_of_Alpha}\\
    A &= \rho(Q,K) \in \mathbb{R} ^ {n \times d} . \label{eq:Alpha_from_rho}
\end{align}
In the above, $q_\text{new}$, $k_\text{new}$ and $v_\text{new}$ are the newest tokens corresponding to the new query, key and value, respectively, $\odot$ denotes a column-aligned element-wise multiplication operation, $\mathbb{1}_n$ is a row-vector of $n$ ones, and $\rho(\Psi, \Omega) = \text{exp}\left(\frac{\Psi \Omega^T}{\sqrt{d}}\right)$. This is in practice causes the decoupling of the numerator and denominator of the softmax activation, allowing its caching into separate operators. We can use Eqs. (\ref{eq:exp_trans})-(\ref{eq:Alpha_from_rho}) to reformulate the SDA continually using the following updates:
\begin{equation}
    \phi(A) = \begin{bmatrix} 
        \phi(\hat{A}) - \rho(Q_\text{mem},k_\text{old}) + \rho(Q_\text{mem},k_\text{new}) \\ \rho(q_{\text{new}},K)\mathbb{1}^T_n
        \end{bmatrix},
\end{equation}
\begin{equation}
    AV = \begin{bmatrix} 
        \hat{AV} - \rho(Q_\text{mem},k_\text{old})v_\text{old} + \rho(Q_\text{mem},k_\text{new})v_\text{new} \\ \rho(q_\text{new},K)V
        \end{bmatrix},
\end{equation}
where the $AV$ matrix is the result of the multiplication between the $A$ and $V$ matrices, $\phi(\hat{A})$ and $\hat{AV}$ are the matrices $\phi(A)$ and $AV$ from the previous update without their first row, respectively, $k_\text{old}$ is the token that shifts out of the attention window when $k_\text{new}$ enters, and $Q_\text{mem}$ are the $n-1$ rows of the query matrix that are still part of the attention window.

In the Continual Retroactive Attention, new tokens update the attention values of all the previous tokens within the attention window. This allows the Continual Transformers to achieve linear computational and memory cost of $O(nd)$. If only the newest token inference is needed, the Single Output Attention can be used to save some additional computations and memory space, i.e.:
\begin{equation}
    \text{Att}_{\text{CoSi}}(q_\text{new}, k_\text{new}, v_\text{new}) = aV \phi(a)^{-1},
\end{equation}
where $a = \rho(q_\text{new},K)$.

\subsection{Nyström-based Attention}
By observing the right-hand side of Eq. (\ref{eq:SDA}), one can make connections of the SDA (specifically its first term) to dot-product formulations appearing in kernel machines \cite{Scholkopf01kernels}. Thus, when the objective is to define an approximate formulation of SDA for large numbers of $n$, matrix approximation schemes like the Nyström approximation \cite{Drineas05nystrom, Wang13nystrom} can be used. This idea was proposed in \cite{Xiong21nystromformer} to define the Nyströmformer model, which approximates the softmax matrix of the SDA in the corresponding attention $\text{Att}_{\text{Ny}}(Q, K, V)$ as follows:
\begin{equation} \label{eq:Nystrom_Attention_Approx}
    \text{s}\left(\frac{QK^T}{\sqrt{d}}\right) \approx \text{s} \left( \frac{Q \tilde{K}^T}{\sqrt{d}} \right) \text{s}\left( \frac{\tilde{Q} \tilde{K}^T}{\sqrt{d}} \right)^\dagger \text{s} \left( \frac{\tilde{Q} K^T}{\sqrt{d}} \right),
\end{equation}
where $\Omega^\dagger$ is the Moore-Penrose pseudo-inverse of matrix $\Omega$, and $\tilde{Q}$ and $\tilde{K}$ are matrices formed by sets of $m$ landmarks, computed as the segment-means of the matrices $Q$ and $K$, respectively. In this context, a landmark is a virtual token aggregates information from $n/m$ real tokens. These landmarks are used to compute the attention in combination with the real tokens in $Q$ and $K$.

When $m\ll n$, it leads to a large reduction of both costs compared to the standard SDA formulation in Eq. (\ref{eq:SDA}), i.e., $O(nd)$ computational and memory costs. In practice, this achieves an approximate attention result that involves the multiplication of 3 matrices of small size that requires less computations than the computation of the original attention, as $\tilde{Q}$ and $\tilde{K}$ have a shape of $m \times d$.

The same formulation has been used in \cite{Patrick21Orthoformer} where, instead of defining the landmarks as the segment-means, landmarks are chosen from the sequence incrementally, in such a way that at each step the most orthogonal vector to the already selected tokens is chosen. In \cite{Dutson23eventful_transformers}, attention is computed over the parts of the image of a video where the most elements have changed, triggering $m$ updates over the image as the rest of the weights are re-used.

\section{Method} \label{sc:me}
In this section, we describe in detail the proposed Continual Transformer model employing a Nyström-based formulation of SDA. Since the SDA approximation in Eq. (\ref{eq:Nystrom_Attention_Approx}) was originally proposed for defining the Nyströmformer model in \cite{Xiong21nystromformer}, we refer to our proposed model as the Continual Nyströmformer. 
The Continual Nyströmformer adapts the computations needed for the Nyström-based SDA in order to be performed in a Continual Inference setting \cite{Hedegaard22continual_inference}.

The first step involves decoupling the softmax operators present in the Nyströmformer attention into their numerator and denominator parts, following \cite{Hedegaard23cont_trans}. As such, we define the Continual Nyström-based SDA as: 
\begin{equation} \label{eq:bgd_att}
\text{Att}_\text{CoNy}(q_\text{new},k_\text{new},v_\text{new}) = \Big(B_\phi (\Gamma_\phi)^\dagger \Delta_\phi \Big) V,
\end{equation}
where for any matrix $\Omega$ (in this case, the matrices $B$, $\Gamma$ and $\Delta$) we have $\Omega_\phi = \phi(\Omega)^{-1} \odot \Omega$. $B$, $\Gamma$ and $\Delta$ are defined as follows:
\begin{align}
    B &= \rho(Q,\tilde{K}) \in \mathbb{R}^{n \times m}, \\
    \Gamma &= \rho(\tilde{Q},\tilde{K}) \in \mathbb{R}^{m \times m}, \\
    \Delta &= \rho(\tilde{Q},K) \in \mathbb{R}^{m \times n}, \label{eq:Delta}
\end{align}
where the matrices $Q$ and $K$ are updated in a continual manner when new tokens are received and $\tilde{Q}$ and $\tilde{K}$ are the landmark matrices used for obtaining the approximation.

As can be seen, in Eqs. (\ref{eq:bgd_att})-(\ref{eq:Delta}), the landmark matrices $\tilde{Q}$ and $\tilde{K}$ are involved in the calculation of in all three matrices $B$, $\Gamma$ and $\Delta$. The Nyströmformer \cite{Xiong21nystromformer} calculates new landmarks for every inference step. However, this approach would lead to computational redundancies in a Continual Inference setting, as landmarks would need to be fully recomputed after every inference step. To address this issue, we exploit properties stemming from the fact that successive inference steps involve processing of highly-overlapping sequence data. We propose two ways for landmark selection, leading to model updates described in Sections \ref{SS:ContLandmarks} and \ref{SS:FixedLandmarks}. Considerations related to the implementation aspects of the proposed model and its updates are discussed in Section \ref{SS:considerations}. Table \ref{tb:costs} shows the asymptotic computational and memory costs of the proposed method compared to the existing ones. A detailed analysis of this can be found in \ref{Appendix:comp} and \ref{Appendix:memory}.

\begin{table*}[hbtp]
    \begin{center}
    \caption{Computational and memory cost of the different attention types. Valley and peak memory costs make reference to the lowest and highest memory costs of each attention type. A detailed analysis can be found in Apendices \ref{Appendix:comp} and \ref{Appendix:memory}.}
    \label{tb:costs}
    \resizebox{1.4\columnwidth}{!}{
        \begin{tabular}{c | c c c} 
            \hline
            \textbf{SDA type} & \textbf{Computational cost} & \textbf{Valley memory cost} & \textbf{Peak memory cost}\\
            \hline \hline
            Att & $O(n^2d)$ & $O(nd)$ & $O(n^2 + nd)$\\
            \hline
            $\text{Att}_\text{Ny}$ & $O(ndm + nm^2 + m^3)$ & $O(nd)$ & $O(nd + nm + m^2)$\\
            \hline
            $\text{Att}_\text{Ny}^\text{Fix}$ & $O(ndm + nm^2)$ & $O(nd + m^2)$ & $O(nd + nm + m^2)$\\
            \hline
            $\text{Att}_\text{CoSi}$ & $O(nd)$ & $O(nd)$ & $O(nd)$\\
            \hline
            $\text{Att}_\text{CoRe}$ & $O(nd)$ & $O(nd)$ & $O(nd)$\\
            \hline
            $\text{Att}_\text{CoNySi}^\text{Cont}$ \textbf{(Ours)} & $O(ndm + m^3)$ & $O(nd + m^2)$ & $O(nd + m^2)$\\
            \hline
            $\text{Att}_\text{CoNyRe}^\text{Cont}$ \textbf{(Ours)} & $O(ndm + nm^2 + m^3)$ & $O(nd + m^2)$ & $O(nd + nm + m^2)$\\
            \hline
            $\text{Att}_\text{CoNySi}^\text{Fix}$ \textbf{(Ours)} & $O(dm + m^2)$ & $O(dm + m^2)$ & $O(dm + m^2)$\\
            \hline
            $\text{Att}_\text{CoNyRe}^\text{Fix}$ \textbf{(Ours)} & $O(ndm + m^2)$ & $O(nm + m^2)$ & $O(nd + nm + m^2)$\\
            \hline
            \hline
        \end{tabular}
    }
    \end{center}
\end{table*}

\begin{figure}[!t]
    \centering
    \includegraphics[width=0.5\textwidth]{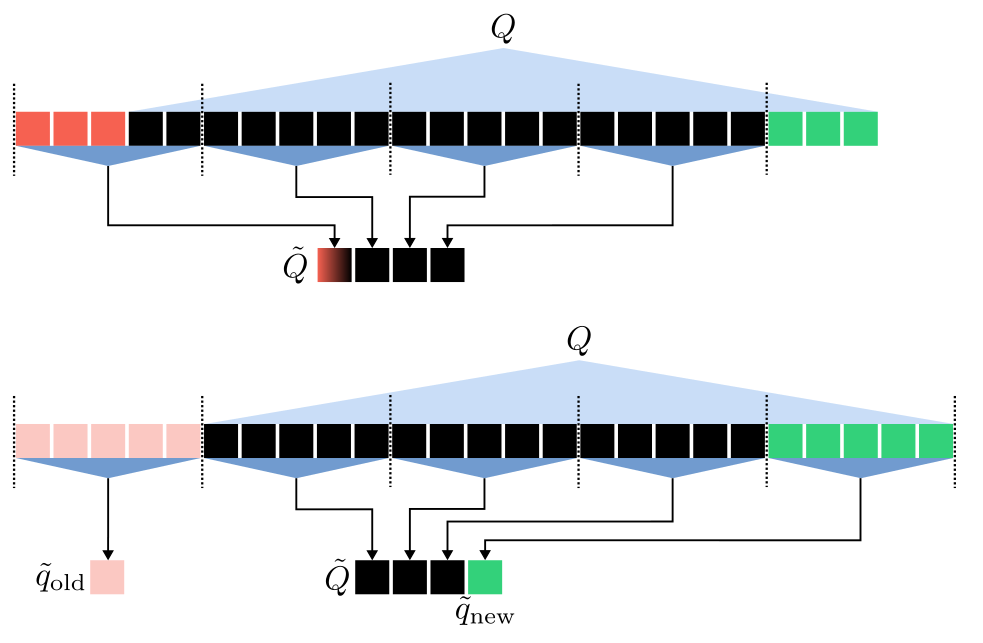}
    \caption{Continual landmarks' calculation (for $n=20$ and $m=4$). Until enough new input tokens have been received, the landmarks remain fixed. This includes the landmark calculated by some \textcolor{red_old}{old} tokens which have been shifted out of the current attention window (top). When enough input tokens have been received (bottom), a new landmark is computed using all the \textcolor{green_new}{new} tokens, replacing the $\tilde{q}_\text{old}$ landmark. The same approach is applied to update $\tilde{K}$.
    \label{fig:continual_landmarks}}
\end{figure}
\begin{figure*}[!t]
    \centering
    \includegraphics[width=0.65\textwidth]{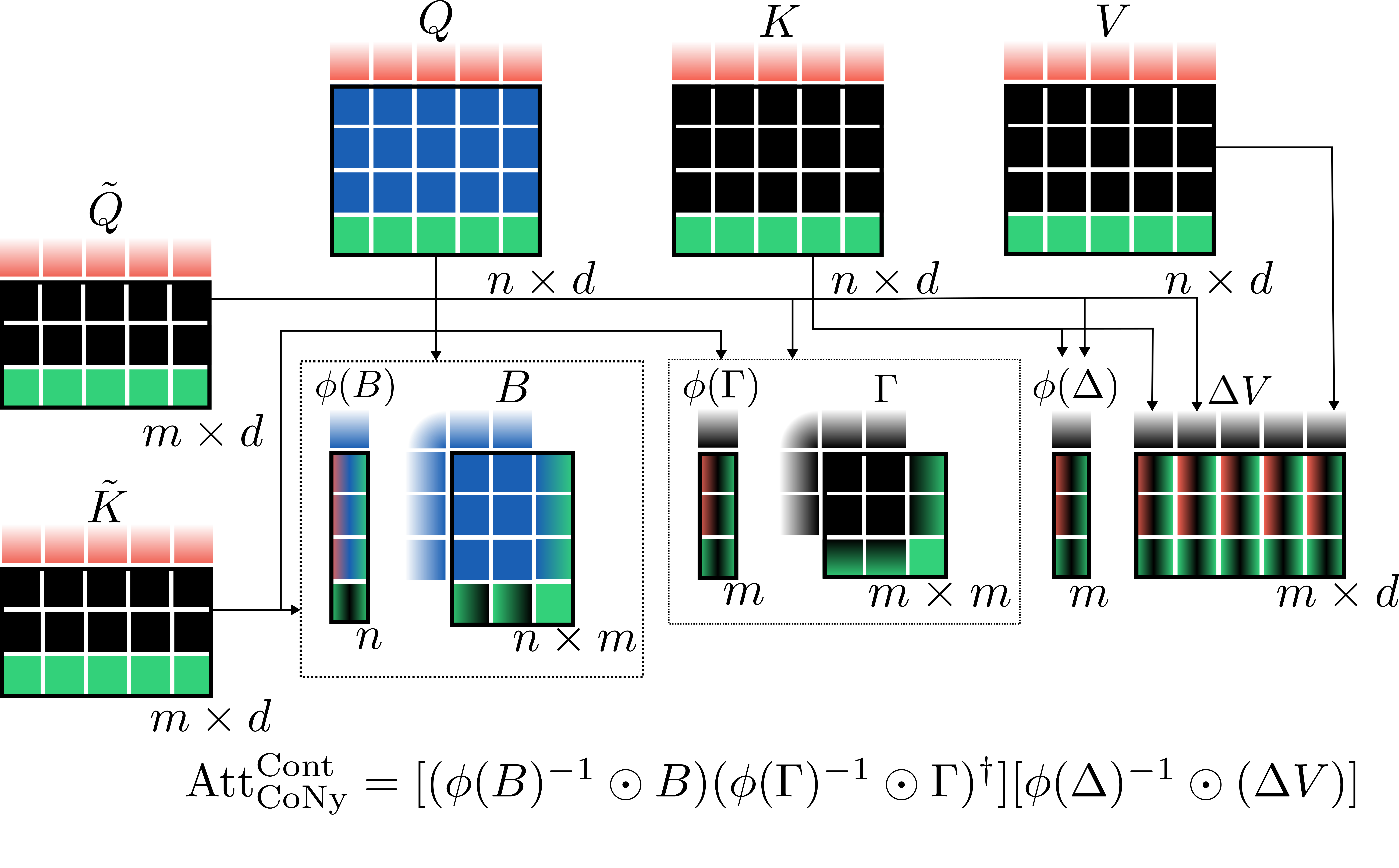}
    \caption{Continual Inference with updated landmarks for a sequence of $n=4$ tokens, each having $d=5$ dimensions, and using $m=3$ landmarks. The \textcolor{red_old}{red} elements represent the tokens that just exited the inference window (old), and the \textcolor{green_new}{green} elements represent the newly received tokens (new). The \textcolor{blue_retro}{blue} elements are those exclusively used by the Retroactive Attention formulation. The elements with a \textcolor{red_old}{red}-black-\textcolor{green_new}{green} color symbolize tokens that need to be updated by removing the influence of the oldest tokens and adding the influence of the newest tokens. This results in updates to all three $B$, $\Gamma$ and $\Delta$ matrices. \label{fig:nystrom_coatt_with_updates}}
\end{figure*}

\subsection{Continual Landmarks}\label{SS:ContLandmarks}
In the Continual Nyströmformers not only the input matrices $Q$ and $K$ get updated, but also their corresponding landmark matrices $\tilde{Q}$ and $\tilde{K}$. We propose updating the landmark matrices after ${n}/{m}$ updates, following the segment-means process used in \cite{Xiong21nystromformer}. We have decided to follow this approach because the segment-means method can also be interpreted as a pooling operation over the temporal dimension. Moreover, this method has been used in existing literature \cite{Lu21softmax_free, Zhang23multi_resolution}.

In practice, this will cause the oldest landmarks $\tilde{q}_\text{old}$ and $\tilde{k}_\text{old}$ (computed by using tokens that have already been shifted out of the $Q$ and $K$ matrices) to be shifted out of the landmark matrices $\tilde{Q}$ and $\tilde{K}$, respectively. This decision allows the model to reuse many of the cached matrices from previous iterations into the computation of future tokens, which allows a more efficient computation in a Continual Inference setting.

The newest landmarks $\tilde{q}_\text{new}$ and $\tilde{k}_\text{new}$, computed as the mean of the most recent ${n}/{m}$ tokens in the $Q$ and $K$ matrices, respectively, will then be included in the landmark matrices $\tilde{Q}$ and $\tilde{K}$. This process can be visualized in Figure \ref{fig:continual_landmarks}. As such, the update of the landmark matrices takes the form:
\begin{align}
    \tilde{Q} = \begin{bmatrix}        \tilde{Q}_\text{mem} \\ \tilde{q}_\text{new} \end{bmatrix},
    & 
    \textrm{where} \:\: \tilde{q}_\text{new} = \frac{m}{n}\sum_{i=n-\frac{n}{m}}^n q_i,\\
    \tilde{K} = \begin{bmatrix}        \tilde{K}_\text{mem} \\ \tilde{k}_\text{new} \end{bmatrix},
    & 
    \textrm{where} \:\: \tilde{k}_\text{new} = \frac{m}{n}\sum_{i=n-\frac{n}{m}}^n k_i,
\end{align}
where $q_i$ and $k_i$ correspond to the $i^\text{th}$ row of matrices $Q$ and $K$, respectively. 

Updating the matrices $B$, $\Gamma$ and $\Delta$ based on continual landmarks takes two forms, depending on whether the landmark matrices have been updated with a new landmark or not, which are described in the following.

\begin{figure*}[!t]
    \centering
    \includegraphics[width=0.65\textwidth]{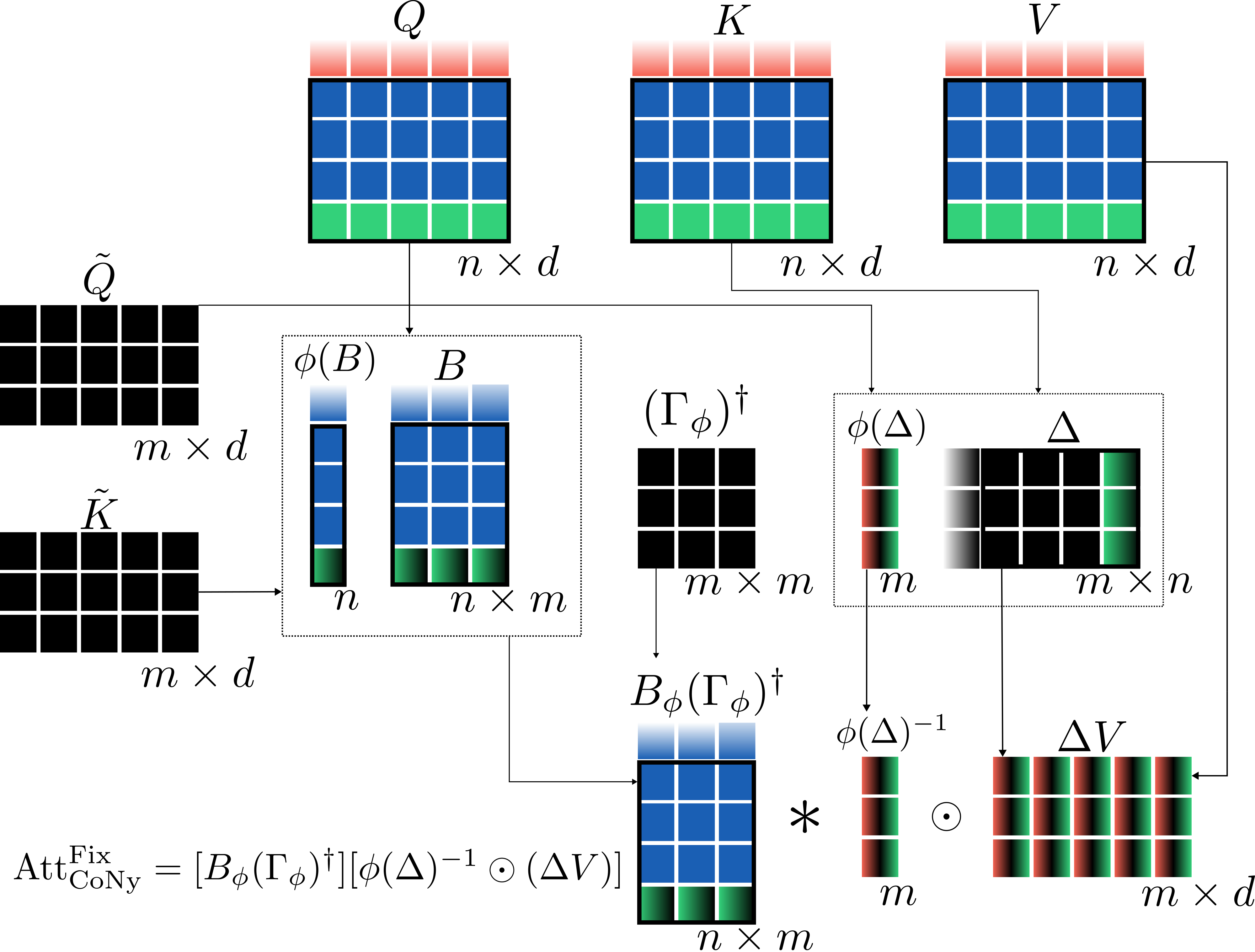}
	\caption{Continual Inference with non-updated landmarks. As landmarks remain unchanged, we can update most of the previous matrices to save computations. The details on color use and dimensionality are identical to those in Figure \ref{fig:nystrom_coatt_with_updates}.\label{fig:nystrom_coatt_no_updates}}
\end{figure*}

\subsubsection{Continual Inference with updated landmarks} \label{SSS:contla}
This is the case where newly received input tokens $q_\text{new}$, $k_\text{new}$ and $v_\text{new}$ lead to the calculation of a new set of landmarks $\tilde{q}_\text{new}$ and $\tilde{k}_\text{new}$, as described above. This means that all matrices $\tilde{Q}$, $\tilde{K}$, $Q$, $K$ and $V$ are continually updated. An illustration of the process followed in this case can be seen in Figure \ref{fig:nystrom_coatt_with_updates}. 

Updating the matrices $B$ and $\Gamma$ in a continual manner given a new token $q_\text{new}$ is done as follows:
\begin{align}
    B &= \begin{bmatrix} B_{\text{mem}} & \rho(Q_\text{mem},\tilde{k}_\text{new})\\
    \multicolumn{2}{c}{\rho(q_\text{new},\tilde{K})} \end{bmatrix},\\
    \Gamma &= \begin{bmatrix} \Gamma_{\text{mem}} & \rho(\tilde{Q}_{\text{mem}},\tilde{k}_\text{new})\\ \multicolumn{2}{c}{\rho(\tilde{q}_\text{new},\tilde{K})} \end{bmatrix}.
\end{align}
Since $\Delta$ has the same form as $B$ and $\Gamma$, a similar update can be used. However, considering that a faster update can be obtained by updating $\Delta$ and $V$ together, we update the matrix $\Delta V$ as follows:
\begin{equation}
    \Delta V= \begin{bmatrix} \hat{\Delta V} - \rho(\tilde{Q}_\text{mem},k_{\text{old}})v_\text{old} + \rho(\tilde{Q}_\text{mem},k_{\text{new}})v_\text{new} \\  \rho(\tilde{q}_\text{new},K)V\end{bmatrix}.
\end{equation}
The matrices $B_\phi$, $(\Gamma_\phi)^\dagger$ and $\Delta_\phi V$ need to be computed from their parts, as all three corresponding vectors $\phi(B)$, $\phi(\Gamma)$, $\phi(\Delta)$ get all their elements changed. This can be expressed as:
\begin{align}
    \phi(B) &= \begin{bmatrix} \phi(\hat{B})-\rho(Q_\text{mem},\tilde{k}_\text{old}) + \rho(Q_\text{mem},\tilde{k}_\text{new}) \\  \rho(q_\text{new},\tilde{K})\mathbb{1}_m^T \end{bmatrix},\\
    \phi(\Gamma) &= \begin{bmatrix} \phi(\hat{\Gamma})-\rho(\tilde{Q}_\text{mem},\tilde{k}_\text{new}) + \rho(\tilde{Q}_\text{mem},\tilde{k}_\text{new}) \\  \rho(\tilde{q}_\text{new},\tilde{K})\mathbb{1}_m^T \end{bmatrix},\\
    \phi(\Delta) &= \begin{bmatrix} \phi(\hat{\Delta})-\rho(\tilde{Q}_\text{mem},k_\text{old}) + \rho(\tilde{Q}_\text{mem},k_\text{new}) \\  \rho(\tilde{q}_\text{new},K)\mathbb{1}_n^T \end{bmatrix}.
\end{align}

The Continual Retroactive Nyströmformer Attention with updated landmarks can then be computed as:
\begin{equation}
    \text{Att}^{\text{ContUp}}_{\text{CoNyRe}} (q_\text{new}, k_\text{new}, v_\text{new}) = \left( B_\phi (\Gamma_\phi)^\dagger\right) \left(\phi(\Delta)^{-1} \odot \Delta V\right),
\end{equation}
where $B_\phi = (\phi(B)^{-1} \odot B)$ and $\Gamma_\phi = (\phi(\Gamma)^{-1} \odot \Gamma)$.

Following the Continual Transformer approach \cite{Hedegaard22Co3DCNN}, a Single Output version of the Continual SDA is also proposed, as there are cases where only the attended features from the newest input token are needed, allowing for further reducing the number of computations. In practice, this is achieved by using the last row of the matrix $B_\phi$ rather than the full matrix i.e.:
\begin{equation}
    \text{Att}^{\text{ContUp}}_{\text{CoNySi}}(q_\text{new}, k_\text{new}, v_\text{new})
    = \left((\beta_\phi)_\text{new} (\Gamma_\phi)^\dagger\right) \left(\phi(\Delta)^{-1} \odot \Delta V\right)
\end{equation}
where $(\beta_\phi)_\text{new}$ is the last row of the $B_\phi$ matrix. This still requires performing the common intermediate computations in the same way as its retroactive counterpart.

\subsubsection{Continual Inference with non-updated landmarks} \label{SSS:FixLa}
This is the case where the newly received input tokens $q_\text{new}$, $k_\text{new}$ and $v_\text{new}$ do not lead to the calculation of a new set of landmarks, thus the matrices $\tilde{Q}$ and $\tilde{K}$ remain identical to those used in the previous inference step. An illustration of the process followed in this case can be seen in Figure \ref{fig:nystrom_coatt_no_updates}. 

We define the following formulation, where updates involve only the new input tokens $q_\text{new}$, $k_\text{new}$ and $v_\text{new}$:
\begin{equation}
    \text{Att}^{\text{ContNUp}}_{\text{CoNyRe}}(q_\text{new}, k_\text{new}, v_\text{new}) =\begin{bmatrix} 
        (B_\phi(\Gamma_\phi)^\dagger)_\text{mem} \\
        \rho(q_\text{new},\tilde{K})_\phi
    \end{bmatrix} \phi(\Delta)^{-1} \odot \Delta V, \label{eq:SDA_fixed_Retro}
\end{equation}
where $(B_\phi(\Gamma_\phi)^\dagger)_\text{mem}$ is the matrix corresponding to the $n-1$ most recent tokens of the matrix $(B_\phi(\Gamma_\phi)^\dagger)$ from the previous iteration. The vector $\phi(\Delta)^{-1}$ and matrix $\Delta V$ require updates to all of their elements, by removing the influence of the oldest tokens and adding the influence of the newest tokens, which is done as follows:
\begin{align}
    \phi(\Delta)^{-1} &= \phi(\Delta)^{-1}_\text{prev}-\rho(\tilde{Q},k_{\text{old}})^{-1} + \rho(\tilde{Q},k_{\text{new}})^{-1}, \\
    \Delta V &=  (\Delta V)_\text{prev} - \rho(\tilde{Q},k_{\text{old}})v_\text{old} + \rho(\tilde{Q},k_{\text{new}})v_\text{new},
\end{align}
where $\phi(\Delta)^{-1}_\text{prev}$ and $(\Delta V)_\text{prev}$ correspond to the matrices obtained in the previous inference step.

Similarly to the continual landmark version, a Single Output simplified version can be formulated, leading to:
\begin{equation}
    \text{Att}^{\text{ContNUp}}_{\text{CoNySi}}(q_\text{new}, k_\text{new}, v_\text{new}) = \left((\beta_\phi)_\text{new}(\Gamma_\phi)^\dagger \right) \phi(\Delta)^{-1} \odot \Delta V. \label{eq:SDA_fixed_Single}
\end{equation}

\subsection{Fixed Landmarks} \label{SS:FixedLandmarks}
The idea of reducing the number of updates can be extended to avoid the landmark computations during inference time. As such, we propose a variant of the Continual Nyströmformers with this feature, avoiding the need to perform landmark updates during Continual Inference.

The most important decision is to determine the appropriate landmarks during the training phase, which can then be used for processing any received input. This approach is motivated by similar ideas used in approximate kernel-based learning \cite{zhang2015scaling,iosifidis2017class}, where the landmarks in Nyström-based approximation of the kernel matrix are redefined as the resulting centroids after clustering the training data. However, this approach cannot be directly applied in our case, as the data transformations performed by all layers before each of the SDA blocks change at every training update, leading to different feature spaces in which the matrices $X_Q$ and $X_K$ are defined.

To address this issue, the training process is divided into two phases. In the first phase, the model is trained in an end-to-end manner using continually updated landmarks as described in Section \ref{SS:ContLandmarks}. The second phase is divided into two processing steps. In the first step, the training data is introduced to the model and the matrices $Q$ and $K$ are calculated for each input data sample. The $Q$-tokens corresponding to all training data are combined to create a dataset which is clustered into $m$ clusters by applying the $m$-Means method. The cluster centers are then used to form the matrix $\tilde{Q}$. The same process is applied to the $K$-tokens to form the matrix $\tilde{K}$. If multiple SDA heads are used, we compute the landmarks of each head independently. We have decided to use cluster centers as the landmarks, as it has been shown that this leads to a better matrix approximation error for the Nyström method \cite{Zhang10clustered_nystrom}.

In the second step, the model is fine-tuned in an end-to-end manner using the now fixed, landmarks (i.e., the matrices $\tilde{Q}$ and $\tilde{K}$ are not updated). When the model is formed by multiple SDA blocks, the two steps of phase two are applied sequentially starting from the first block, and keeping all landmarks of previous SDA blocks fixed in the fine-tuning step. This leads to gradually determining all landmarks of the model. 

After training the model and determining all landmarks, the SDA module used for Continual Retroactive Inference, i.e., $\text{Att}_\text{CoNyRe}^\text{Fix}$, has the form of Eq. (\ref{eq:SDA_fixed_Retro}), and the SDA module for the Continual Single Output Inference, i.e., $\text{Att}_\text{CoNySi}^\text{Fix}$, has the form of Eq. (\ref{eq:SDA_fixed_Single}). 
The computational cost of this model is identical to the Nyström-based Continual Inference with non-updated landmarks for both the Retroactive and Single Output versions.

\subsection{Implementation Aspects} \label{SS:considerations}
The Continual Nyströmformers share some of the practical aspects of Continual Transformers \cite{Hedegaard23cont_trans}, due to the properties of the involved continual computations:
\begin{itemize}
    \item The Continual Nyströmformers require a circular positional encoding, as when new input tokens are processed its positional encoding needs to be appropriately related to the positional encodings of the rest of the sequence.
    \item The ability for Continual Nyströmformers to reuse previous computations is hampered when multiple stacked SDA blocks are used. This is caused by the need to recompute the entire sequence for all the earliest SDA blocks, as the attention needs to be propagated accordingly. Thus, all SDA blocks except the last one must be of a regular Nyströmformer or any other non-Continual Transformer.
    \item For training, we use a modified version of the non-continual model with the circular positional encoding described above and the corresponding landmark selection scheme as described in Sections \ref{SS:ContLandmarks} and \ref{SS:FixedLandmarks}, depending on whether continual or fixed landmarks are used, respectively. We follow this approach as the non-continual training processes are faster when the entire sequence is available from the beginning, and both continual and non-continual SDA variants produce identical results.
\end{itemize}

Some aspects that affect the implementation of the continual landmarks and fixed landmarks selection schemes are:
\begin{itemize}
    \item For the calculation of the continual landmarks in Section \ref{SS:ContLandmarks}, even though $n$ and $m$ are hyperparameters which can be chosen by the user such that $(n \bmod m) = 0$, in the general case where this condition is not met, the following process is used. The first $(n \bmod m)$ landmarks are calculated by using a segment of the token sequence that has an extra token. The position of these landmarks is tracked as newer landmarks are included and older landmarks are discarded, so every new landmark will be calculated using a segment of the token sequence of the same size as the landmark it is replacing.

    \item For the calculation of the fixed landmarks in Section \ref{SS:FixedLandmarks}, when the size of the training set is prohibitive to be used for the formation of the $Q$-token and $K$-token datasets for the $m$-Means clustering, a (randomly-chosen) subset of the training data can be used instead. 
    
    \item For the calculation of the fixed landmarks in Section \ref{SS:FixedLandmarks}, in the case where $X_Q = X_K$, i.e., when SDA is used for self-attention, one can determine the fixed landmarks by performing clustering once, e.g., on the $X_Q$ dataset to determine cluster centers $\tilde{X}_Q$, and then calculate the two landmark sets $\tilde{Q} = \tilde{X}_Q W_Q$ and $\tilde{K} = \tilde{X}_K W_K$.

    \item For the Continual Inference with updated landmarks process in Section \ref{SSS:contla}, the update of matrix $\Gamma_\phi$ requires a full re-computation of its Moore-Penrose pseudo-inverse since, to the best of our knowledge, there is no existing method to compute the Moore-Penrose pseudo-inverse continually. Computing exactly the pseudo-inverse involves costly Singular Value Decomposition operations. Because of this, we adopt the approach used in \cite{Xiong21nystromformer} relying on the iterative method from \cite{Razavi14pinv} to obtain the pseudo-inverse.

    \item When a Continual Inference step does not involve updating of the landmark matrices $\tilde{Q}$ and $\tilde{K}$, as it is the case described in Sections \ref{SSS:FixLa} and \ref{SS:FixedLandmarks}, the calculation of the matrix $\Gamma_\phi$ and its Moore-Penrose pseudo-inverse is not needed, as it has been already done in the previous inference step for the process described in Section \ref{SSS:FixLa}, or it has been calculated and stored in the training phase for the process described in Section \ref{SS:FixedLandmarks}.

    \item Even though the updates for the proposed Continual Nyström-based SDA formulations are provided for single new tokens $q_\text{new}$, $k_\text{new}$, and $v_\text{new}$, their formulation for multiple token-based updates (e.g., incorporating the tokens corresponding to the multiple patches comprising a new video frame when processing a visual stream) can be obtained in a straightforward manner by using $Q_\text{new}$, $K_\text{new}$, and $V_\text{new}$ instead.
\end{itemize}

For replacing the softmax with exponential operations in both the Continual Transformers and our Continual Nyströmformers numerical stability issues need to be considered, as the calculation of the exponential can be prone to overflow and underflow. This is addressed in most softmax implementations (e.g., \cite{Paszke19Pytorch}) by employing the so-called stable softmax variant \cite{Goodfellow16DL} $\text{s}(x)_i = \text{exp}(x_i - C) / \sum_{j=1}^n \text{exp}(x_j - C)$. By setting $C$ to the maximum value in $x$ possible overflow and underflow issues are addressed, as at least one value will be higher than zero after the calculation of the exponential operations.
However, this approach cannot be applied in the continual versions of SDA, as the maximum values of $x$ in the attention window can change at every inference step. This would cause constant updates in most matrices, increasing the cost very significantly. In our experiments described in the next section, this issue has not been observed. In case such a stability issue is observed, using a dropout layer or other type of normalization layer before the SDA module can address it.

\begin{figure}[!t]
\centering
\subfloat[]{\includegraphics[width=0.24\textwidth]{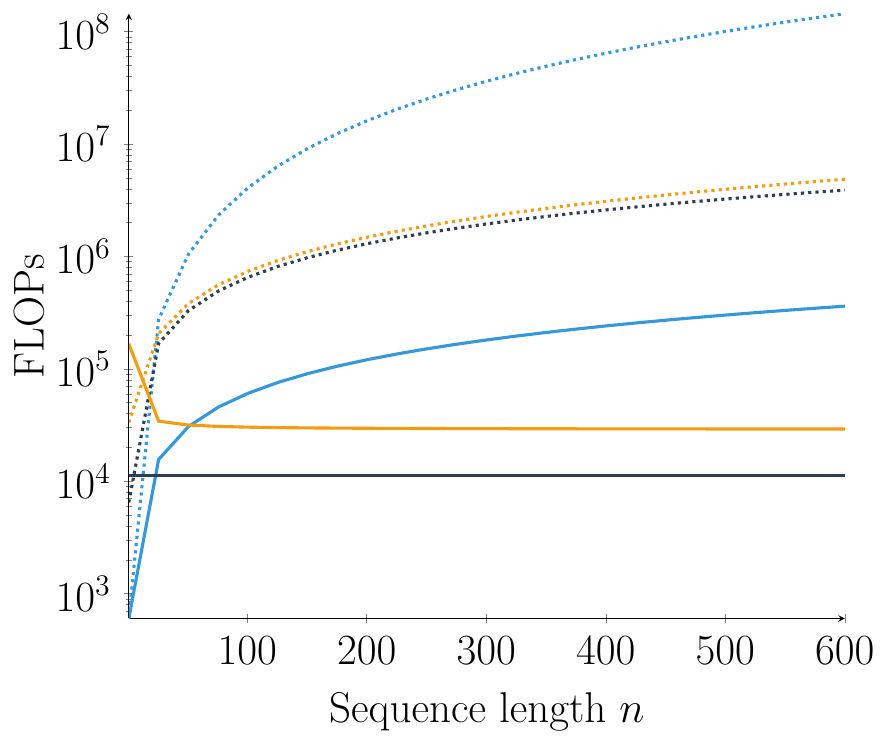}}
\hfil
\subfloat[]{\includegraphics[width=0.24\textwidth]{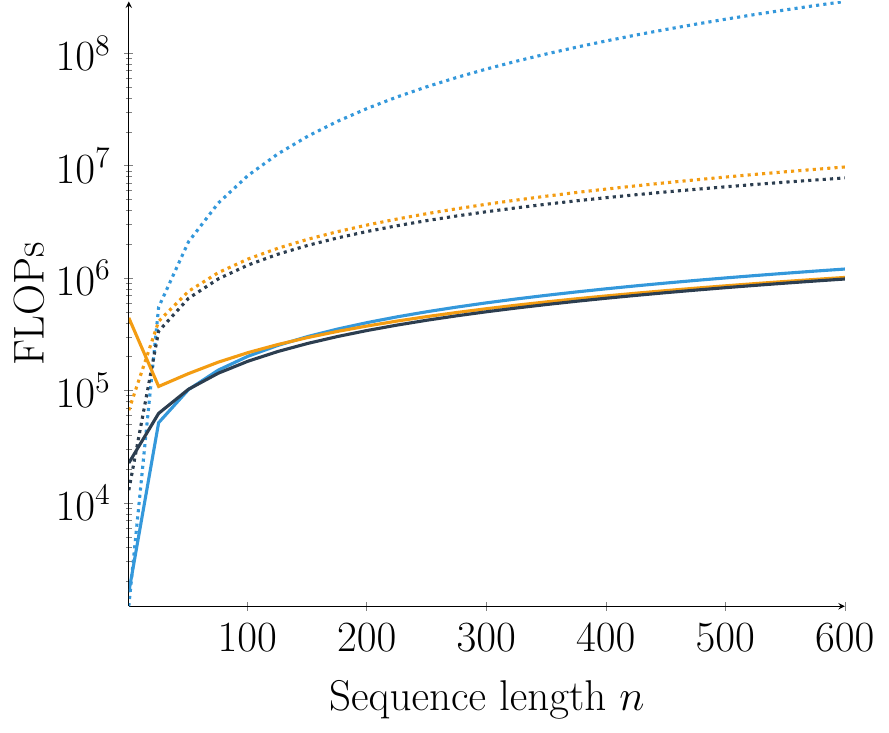}}
\\
\subfloat[]{\includegraphics[width=0.24\textwidth]{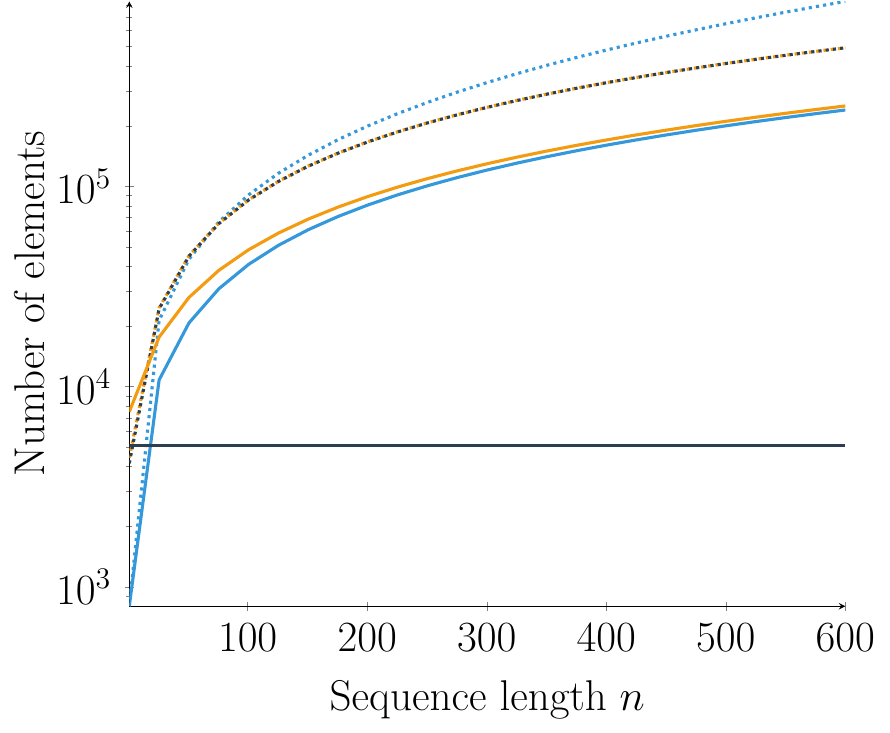}}
\hfil
\subfloat[]{\includegraphics[width=0.24\textwidth]{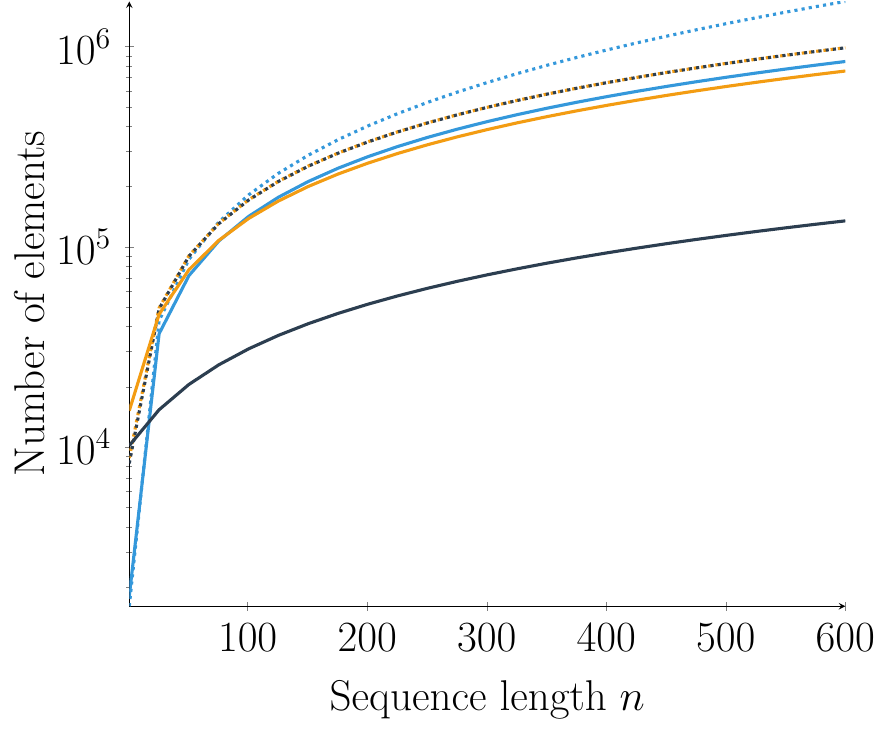}}
\\
\subfloat{\includegraphics[width=6cm]{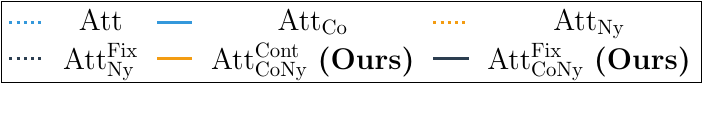}}
\caption{(a) and (b) contain a comparison of computational cost in FLOPs between the SDA modules of different methods in relation to the number of tokens $n$ for 1 and 2 layers, respectively. (c) and (d) contain a comparison of the peak memory between the SDA modules of different methods. Both methods use $d=200$ and $m=8$.}
\label{fig:flop_mem_cost}
\end{figure}

\section{Experiments} \label{sc:ex}
In this section, we provide comparisons of models using the proposed SDA formulations with models using the original SDA formulation \cite{Vaswani17transformer}, the Nyström-based SDA formulation \cite{Xiong21nystromformer} and the Continual SDA formulation \cite{Hedegaard23cont_trans} in terms of their memory and computational costs, as well as their performance on Audio Classification and Online Action Detection tasks. A summary of the properties of these models is provided in Table~\ref{tb:models}.

\begin{table}[hbtp]
    \caption{Properties of the SDA modules used in our experiments.}
    \label{tb:models}
    \begin{center}
        \begin{tabular}{c | c c c } 
            \hline
            \textbf{SDA type} & \textbf{Continual} & \textbf{Nyström-based} & \textbf{Landmark type}\\
            \hline \hline
            Att &  &  & - \\
            \hline
            $\text{Att}_\text{Co}$ & $\checkmark$ &  & - \\
            \hline
            $\text{Att}_\text{Ny}$ &  & $\checkmark$ & Segment-means \\
            \hline
            $\text{Att}_\text{Ny}^\text{Fix}$ &  & $\checkmark$ & Fixed \\
            \hline
            $\text{Att}_\text{CoNy}^\text{Cont}$ \textbf{(Ours)} & $\checkmark$ & $\checkmark$ & Continual \\
            \hline
            $\text{Att}_\text{CoNy}^\text{Fix}$ \textbf{(Ours)} & $\checkmark$ & $\checkmark$ & Fixed \\
            \hline
            \hline
        \end{tabular}
    \end{center}
\end{table}

\subsection{Computational cost experiments}
There exist multiple ways to count and parametrize the actual computational cost of running a model for inference. A metric that is extensively used is the number of floating operations (FLOPs) which corresponds to the number of element computations required to perform inference. To study the computational efficiency of the proposed SDA formulation in comparison with other related formulations for different lengths of input sequences $n$, we provide the number of FLOPs for different sequence lengths when using a number of dimensions $d = 200$ and $m=8$ landmarks in Figure \ref{fig:flop_mem_cost}. As the proposed method affects exclusively the SDA, the figure illustrates the number of average FLOPs required for a single prediction during sequential processing corresponding to the SDA modules of the competing methods. 

Since the proposed approach of determining the landmarks in the training phase and fixing them for performing inference for any input sample in the test phase can be used also by the Nyströmformer model, we also created a variant of the Nyströmformer using fixed landmarks and illustrate its computational cost as $\text{Att}^{\text{Fix}}_{\text{Ny}}$ in Figure \ref{fig:flop_mem_cost}. For the Continual Inference models using one SDA block their single output versions are used, while for those using two SDA blocks an SDA with retroactive inference is followed by a single output SDA. 

The first aspect to notice is that the original SDA formulation $\text{Att}$ leads to the highest computational cost, which has a quadratic asymptotic form. The SDA formulations $\text{Att}_{\text{Ny}}$ and $\text{Att}_{\text{Co}}$ used by the Nyströmformer and Continual Transformer models have a similar asymptotic form for their computational cost as the original SDA formulation, but with much lower computations compared to the original SDA formulation, while the Nyströmformer has a higher number of computations than the Continual Transformer. 

For the proposed SDA formulations $\text{Att}^{\text{Cont}}_{\text{CoNy}}$ and  $\text{Att}^{\text{Fix}}_{\text{CoNy}}$ we need to distinguish two cases, i.e., when one or two SDA blocks are used. When two SDA blocks are used, the computational cost of the proposed SDA formulations has a very similar form to that of the SDA formulation $\text{Att}_{\text{Co}}$ used in the Continual Transformer, while being lower in absolute numbers. The difference between these computational costs depends on the selected number of landmarks $m$ and, as the number of sequence tokens $n$ increases, the difference in the computational costs between the two types of SDA formulations increases.

When one SDA block is used, the computational cost of the proposed SDA formulation $\text{Att}^{\text{Fix}}_{\text{CoNy}}$ using fixed landmarks is not dependent on the number of tokens $n$ and it is the lowest compared to all competing cases. The computational cost of the proposed SDA formulation $\text{Att}^{\text{Cont}}_{\text{CoNy}}$ is higher for smaller number of tokens $n$ as, when the same number of landmarks $m$ is used, landmarks updates take place more frequently. This same behavior can be seen when two layers are used. As the number of tokens $n$ increases, the frequency of landmark updates becomes lower, leading to lower computational cost. For very large numbers of tokens $n$, the computational cost of the proposed SDA formulation $\text{Att}^{\text{Cont}}_{\text{CoNy}}$ becomes (almost) independent to $n$ reaching asymptotically the computational cost of the proposed SDA formulation $\text{Att}^{\text{Fix}}_{\text{CoNy}}$, as the additional computations needed for landmark updates are added every $n$ updates.

\subsection{Memory overhead experiments}
When measuring the memory overhead of a model, we need to consider not only the matrices we need to store between iterations, but also the necessary matrices to perform the computations leading to the output. With this in mind, we provided memory costs of the different SDA formulations based on the dimensions of the input, the sequence length and the number of landmarks in Section \ref{sc:me}. We used these to measure the peak memory overhead of the SDA module with $d=200$ and $m=8$ for a varying sequence length $n$, as shown in Figure \ref{fig:flop_mem_cost}.

Similar to the computational cost, the memory cost of the original SDA formulation $\text{Att}$ is the biggest due to its quadratic form. On the other hand, the proposed SDA formulation $\text{Att}^{\text{Fix}}_{\text{CoNy}}$ using fixed landmarks exhibits the lowest cost, as it does not require storing any of the previously received tokens, and the only memory matrix that requires storing that depends on the sequence length $n$ is the final attention product in the Retroactive Attention version.

The rest of the SDA formulations exhibit a similar behavior, as all of them require storing intermediate matrices with size $n \times d$, including the full matrices $Q$, $K$ and $V$. The SDA formulations $\text{Att}_{\text{CoNy}}$ and $\text{Att}^{\text{Fix}}_{\text{CoNy}}$ have the highest memory cost in this group because they require storing the same set of matrices at peak memory time. 
The SDA formulations $\text{Att}_{\text{Co}}$ and $\text{Att}^{\text{Cont}}_{\text{CoNy}}$ have a very similar memory costs. This is due to that both SDA formulations need to store the same number of matrices of size $n \times d$ at peak memory time.

\begin{table}[hbtp]
    \caption{Audio classification on the GTZAN dataset.\label{table:audio_results}}
    \begin{center}
    \resizebox{\columnwidth}{!}{
        \begin{tabular}{c c | c c c} 
            \hline
            \textbf{Model} & \textbf{Layers} & \textbf{Acc. (\%)} & \textbf{FLOPs (M)} & \textbf{Rel. FLOPs}\\
            \hline \hline
            \multirow{2}{*}{Att} & 1& $93.12\pm5.44$ & 5.57 & $\times$1 \\
             & 2& $94.32\pm4.31$ & 11.13 & $\times$1 \\
             \hline
            \multirow{2}{*}{$\text{Att}_\text{Co}$} & 1& $93.16\pm5.39$ & 0.07 & $\times$80.26\\
            & 2& $94.52\pm4.05$ & 0.39 & $\times$48.26\\
            \hline
            \multirow{2}{*}{$\text{Att}_\text{Ny}$-4}  & 1& 93.54$\pm4.26$ & 0.42 & $\times$13.17\\
            & 2& 93.92$\pm4.43$ & 0.85 & $\times$13.17\\
            \hline
            \multirow{2}{*}{$\text{Att}_\text{Ny}^\text{Fix}$-4}  & 1& $93.24\pm5.30$ & 0.37 & $\times$14.98\\
            & 2& $93.76\pm5.14$ & 0.74 & $\times$14.98\\
            \hline \hline
            \multirow{2}{*}{$\text{Att}_\text{CoNy}^\text{Cont}$-4 \textbf{(Ours)}} & 1& $92.92\pm5.43$ & 0.01 & $\times$509.66 \\
            & 2& $93.54\pm4.91$ & 0.11 & $\times$102.57 \\
            \hline
            \multirow{2}{*}{$\text{Att}_\text{CoNy}^\text{Fix}$-4 \textbf{(Ours)}} & 1& $90.76\pm5.52$ & \textbf{0.005} & \textbf{$\times$1028}\\
            & 2& $92.36\pm5.52$ & 0.10 & $\times$108.92 \\
            \hline
        \end{tabular}
    }
    \end{center}
\end{table}

\begin{table*}[hbtp]
    \caption{Online Action Detection on the THUMOS14 dataset using a Temporal Segment Network pre-trained on the Kinetics-400 dataset or ActivityNet dataset. \label{table:cooadtr}}
    \begin{center}
        \begin{tabular}{c c | c c | c c | c c} 
            \hline
             & & \multicolumn{2}{c|}{\textbf{Kinetics-400}} & \multicolumn{2}{c|}{\textbf{ActivityNet}} & \\
            \textbf{Model} & \textbf{Layers} & \textbf{mAP (\%)} & \textbf{cmAP (\%)} & \textbf{mAP (\%)} & \textbf{cmAP (\%)} & \textbf{FLOPs (M)} & \textbf{Rel. FLOPs}\\
            \hline \hline
            \multirow{2}{*}{Att} & 1 & $64.32\pm0.66$ & $98.27\pm0.05$ & $56.00\pm0.71$ & $97.39\pm0.04$ & 8.46 & $\times$1\\
            & 2 & $\textbf{64.66}\pm0.30$ & $98.28\pm0.06$ & $\textbf{56.95}\pm0.80$ & $97.36\pm0.06$ & 16.92 & $\times$1\\
            \hline
            \multirow{2}{*}{$\text{Att}_\text{Co}$} & 1 & $63.89\pm0.14$ & $98.33\pm0.05$ & $55.69\pm0.41$ & $\textbf{97.43}\pm0.04$ & 0.20 & $\times$42.99\\
            & 2 & $63.93\pm0.43$ & $\textbf{98.34}\pm0.02$ & $56.04\pm0.27$ & $97.40\pm0.08$ & 0.65 & $\times$25.88\\
            \hline
            \multirow{2}{*}{$\text{Att}_\text{Ny}$-16}& 1 & $59.74\pm0.29$ & $97.72\pm0.03$ & $50.57\pm0.52$ & $96.63\pm0.05$ & 4.71 & $\times$1.80\\
            & 2 & $59.32\pm0.23$ & $97.72\pm0.04$ & $50.24\pm0.14$ & $96.47\pm0.03$ & 9.42 & $\times$1.80\\
            \hline
            \multirow{2}{*}{$\text{Att}_\text{Ny}^\text{Fix}$-16} & 1 & $56.77\pm0.51$ & $97.46\pm0.11$ & $48.80\pm0.75$ & $96.41\pm0.12$ & 4.21 & $\times$2.01\\
            & 2 & $53.80\pm2.19$ & $97.28\pm0.29$ & $48.53\pm0.61$ & $96.19\pm0.11$ & 8.43 & $\times$2.01\\
            \hline
            \hline
            \multirow{2}{*}{$\text{Att}_\text{CoNy}^\text{Cont}$-16 \textbf{(Ours)}} & 1 & $60.03\pm0.33$ & $97.70\pm0.03$ & $50.60\pm0.13$ & $96.64\pm0.03$ & 0.46 & $\times$18.35\\
            & 2 & $59.30\pm0.39$ & $97.64\pm0.07$ & $50.68\pm0.38$ & $96.64\pm0.04$ & 1.43 & $\times$11.84\\
            \hline
            \multirow{2}{*}{$\text{Att}_\text{CoNy}^\text{Fix}$-16 \textbf{(Ours)}} & 1 & $55.48\pm0.26$ & $97.21\pm0.07$ &  $47.92\pm0.69$ & $96.05\pm0.05$ & \textbf{0.12} & $\times$\textbf{73.52}\\
            & 2 & $52.41\pm0.65$ & $96.82\pm0.13$ & $46.60\pm0.36$ & $95.76\pm0.06$ & 1.26 & $\times$13.40\\
            \hline
        \end{tabular}
    \end{center}
\end{table*}

\begin{figure}[!t]
\centering
\vtop{
\includegraphics[width=0.48\textwidth]{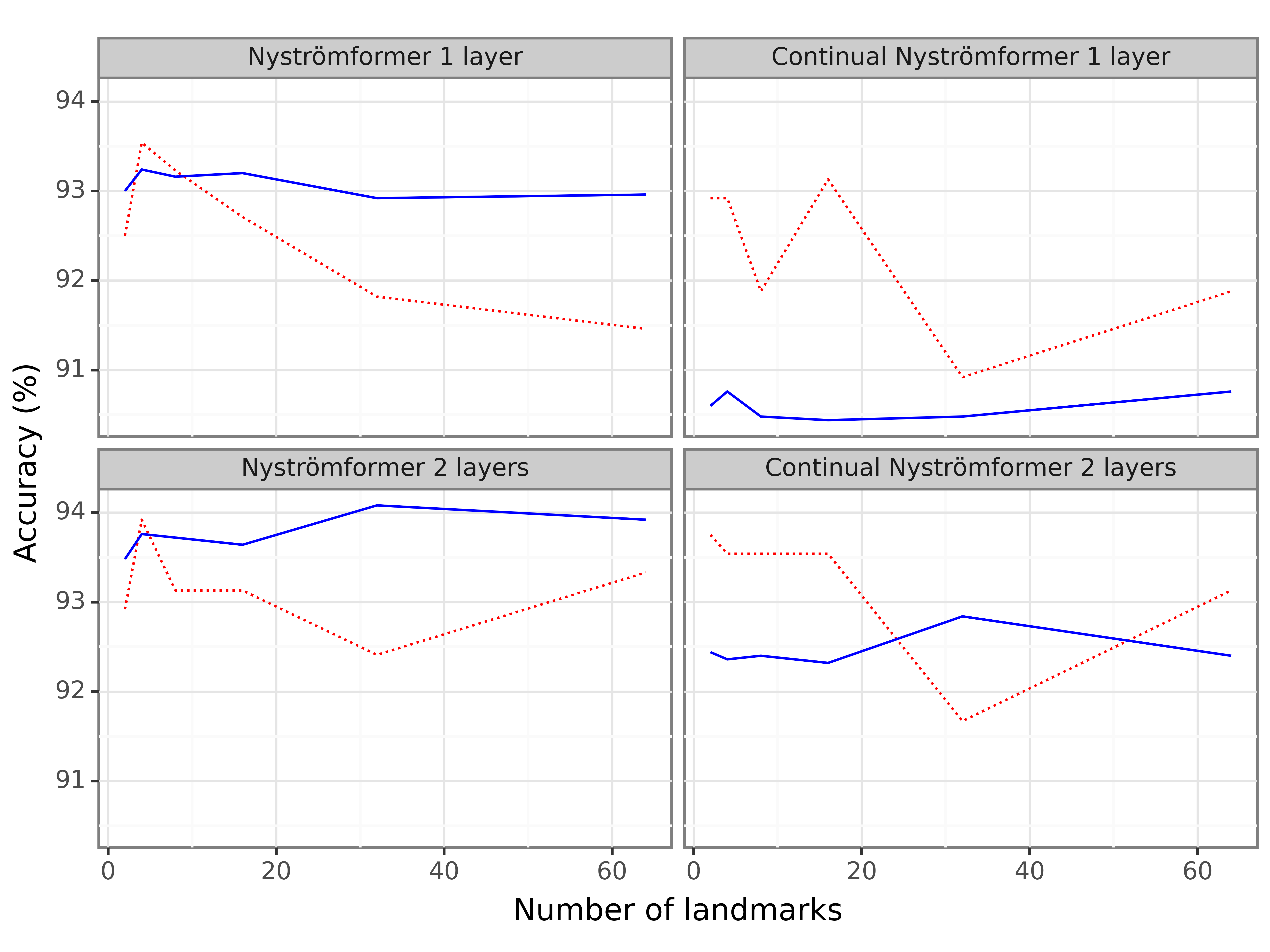}
\caption{Accuracy on the GTZAN dataset of the models using Nyström-based SDA formulations for a varying number of landmarks. \textcolor{blue}{\underline{Blue lines}} refer to the models using fixed landmarks and \textcolor{red}{\dotuline{red lines}} refer to the models either re-calculating landmarks or updating them continually.\label{fig:gtzan_landmarks}}
}
\end{figure}

\subsection{Experiments on real data}

\subsubsection{Audio Classification experiments}
Audio Classification can be defined as the task of labeling an audio track into one or more categories. A common way to process the audio signal is to calculate its Mel Spectrograms \cite{Stevens37mel}, transforming the audio signal into images highlighting properties of the signal that have been shown to be important in human hearing \cite{Choi16melhuman}. Then, these spectrograms can be processed as regular images \cite{Palanisamy20audiocnn}.

In our experiments, we used the GTZAN Music Genre Classification dataset \cite{Tzanetakis02GTZAN}, which contains 100 music tracks for each of the 10 different music genres. Each track has a length of 30 seconds. Following the same architecture as previous works \cite{Hedegaard23cont_trans}, we generate the Mel Spectrogram of each track. This spectrogram is an image where the width and height are the temporal and feature dimensions, respectively. Then, we divide it into one-second clips with a stride of 0.25 seconds, resulting in 120 clips. These are then introduced to a VGGish network \cite{Simonyan15vgg, Hershey17AudioSet} pre-trained on the AudioSet dataset \cite{Hershey17AudioSet}, resulting in a sequence length ($n$) of 120 tokens ready to be used as the input to a Transformer model.

For training, we input the full sequences of 120 tokens as a single attention window into the Transformer-based models and train them with the single label assigned to each track. For testing, non-Continual Inference models receive the full sequences of 120 tokens to provide their response. Continual Inference models have the capacity of computing inference reusing prior operations, as the process described above resembles a continual audio classification process converting the audio signal captured from a microphone in real time into a token every 0.25 seconds, using this token and the previously received ones to perform Continual Inference. 

We performed the experiments with five different data splits between the training, validation and test sets and five different random weight initializations, resulting in 25 runs. For each of the competing methods, we evaluated two models with one and two Transformer layers, respectively, each having 16 heads in the SDA module and token dimensionality of $d = 192$. Table~\ref{table:audio_results} provides the results obtained by using models having the competing SDA formulations. In the table, FLOPs are counted in millions, the last column references the relative amount of FLOPs of each SDA in relation to the original SDA with the same number of Transformer layers, and all Nyström-based models in that table use $m=4$ landmarks. The results in accuracy are similar in general terms. The models with the original SDA and Continual SDA formulations achieve a slightly higher accuracy compared to the models having a Nyström-based SDA formulation. Using two Transformer layers results in a slight increase in accuracy. As for the number of computations, the continual models are significantly more efficient, with the model using the proposed SDA formulation with fixed landmarks having the lowest computational cost.

Figure \ref{fig:gtzan_landmarks} illustrates the classification accuracy obtained by the original Nyströmformer model, its variant using fixed landmarks, and the proposed Continual models using continually updated and fixed landmarks for different number of landmarks in the set $m = \{ 2, 4, 8, 16, 32, 64\}$. Interestingly, the new variant of the Nyströmformer model using fixed landmarks consistently outperforms the original Nyströmformer. On the other hand, the continual landmarks generally offer higher accuracy compared to using fixed landmarks for the proposed Continual Nyströmformers. The accuracy with few landmarks is slightly higher in all cases, but no clear trend can be observed.

\begin{figure}[!t]
\centering
\vtop{
\includegraphics[width=0.48\textwidth]{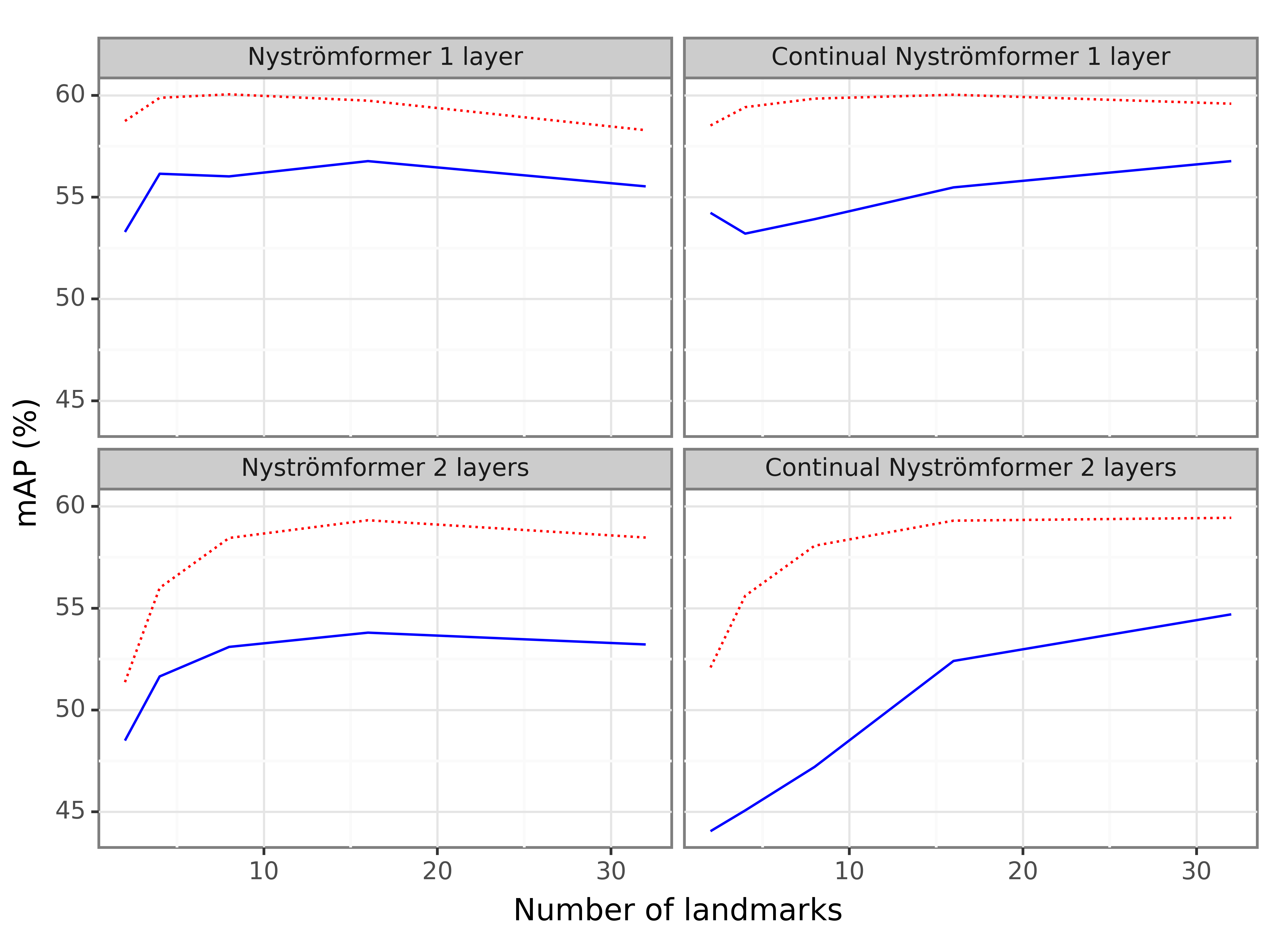}
\caption{mAP on the THUMOS14 dataset of the models using Nyström-based SDA formulations for a varying number of landmarks, when the models are pre-trained on the Kinetics-400 dataset. \textcolor{blue}{\underline{Blue lines}} refer to the models using fixed landmarks and \textcolor{red}{\dotuline{red lines}} refer to the models either re-calculating landmarks or updating them continually.\label{fig:oadkin_landmarks}}
}
\end{figure}
\begin{figure}[!t]
\centering
\vtop{
\includegraphics[width=0.48\textwidth]{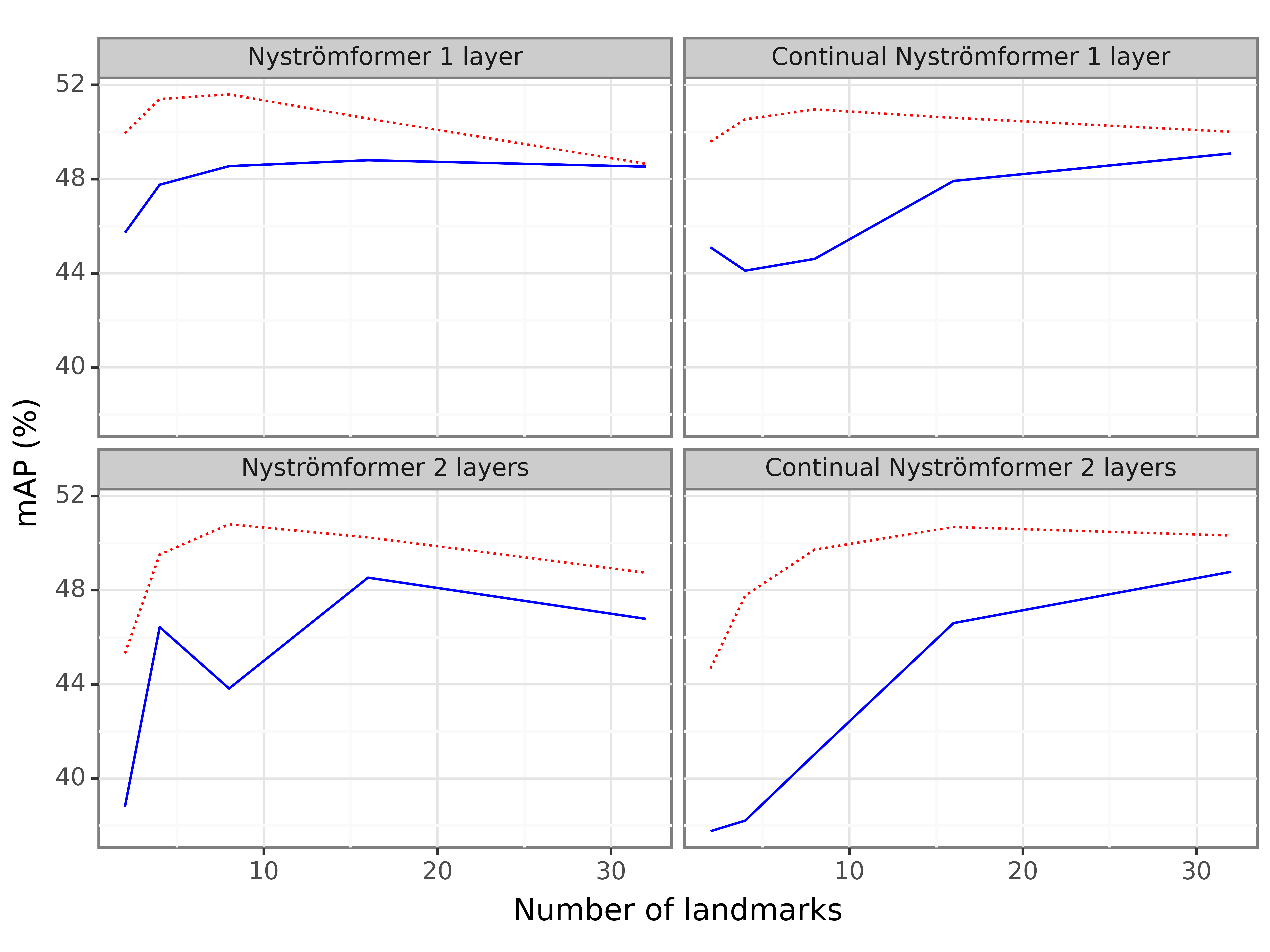}
\caption{mAP on the THUMOS14 dataset of the models using Nyström-based SDA formulations for a varying number of landmarks, when the models are pre-trained on the ActivityNet dataset. \textcolor{blue}{\underline{Blue lines}} refer to the models using fixed landmarks and \textcolor{red}{\dotuline{red lines}} refer to the models either re-calculating landmarks or updating them continually.\label{fig:oadanet_landmarks}}
}
\end{figure}

\subsubsection{Online Action Detection experiments}
Online Action Detection \cite{DeGeest16OAD} is the task of detecting actions in videos, i.e., the detection of the starting video frame of the action and its classification to a number of known action classes, without considering future information. Such an experimental setup imitates the Continual Inference setting of real-life applications, which is well-suited for Continual Inference models.

In our experiments, we used the THUMOS14 dataset \cite{Idrees17THUMOS}, formed by 413 videos which have frame-level labeling for 20 action classes. Following the same approach as previous works \cite{Hedegaard23cont_trans, Lee23info_OAD}, we use abstract features extracted using a Temporal Segment Network \cite{Wang19temporal_segment_networks}, which has been previously trained over the ActivityNet dataset \cite{Heilbron15ActivityNet} or Kinetics-400 dataset \cite{Carreira17Kinetics}. Similar to the Audio Classification task, all models are trained on video clips of 64 video frames, non-Continual models are evaluated in the test phase using video clips of the same size as in the training phase, i.e., 64 video frames, and Continual models perform inference in a continual manner by updating their attention window using the tokens of the newly collected video frames.

The metrics used to evaluate the performance of the models are the mean Average Precision (mAP) and the calibrated mean Average Precision (cmAP) \cite{DeGeest16OAD, Lee23info_OAD}. The same architecture is used as in \cite{Hedegaard23cont_trans}, and all Nyström-based models use 16 landmarks. For the Nyström-based models that use fixed landmarks, we use a subset of the training video tokens to perform landmark selection, i.e., we randomly select 50,000 tokens for performing $m$-Means clustering. Following the experimental protocol for Audio Classification, five different models and five data seeds are used.

Table~\ref{table:cooadtr} provides the results obtained by using the Segment Network model pre-trained on the Kinetics-400 \cite{Carreira17Kinetics} and the ActivityNet \cite{Heilbron15ActivityNet} datasets, respectively. As can be seen in these tables, models using Nyström-based SDA formulations achieve lower mAP values compared to the models using the original SDA and the Continual SDA formulations, in exchange for a reduced computational cost. The use of fixed landmarks models translates in another reduction in mAP. In terms of cmAP, we can see that the performance of all models is very similar. No significant difference in performance is observed between the models using continual SDA formulations and their non-continual counterparts.

Figures \ref{fig:oadkin_landmarks} and \ref{fig:oadanet_landmarks} illustrate the precision (mAP) obtained by the original Nyströmformer model, its variant using the proposed process for determining landmarks in the training phase and fixing them for inference, and the proposed Continual models using continually updated and fixed landmarks for different number of landmarks in the set $m = \{2, 4, 8, 16, 32\}$. As can be seen, the models using updated landmarks consistently lead to higher precision values compared to their counterparts using fixed landmarks. In these cases, a higher number of landmarks (16 or 32 landmarks) offer higher precision rates, between 2 and 5\%.

\begin{table}[hbtp]
    \begin{center}
        \caption{Electricity usage forecasting results on the Electricity Load Diagrams dataset. All the models use a single transformer layer, and all the Nyström-based models use 4 landmarks.\label{table:electricity}}
        \begin{tabular}{c | c c c} 
            \hline
            \textbf{Model} & \textbf{MSE (10e-3)} & \textbf{FLOPs (M)} & \textbf{Rel. FLOPs}\\
            \hline \hline
            \multirow{1}{*}{Att} & $9.69\pm1.10$ & 67.70 & $\times$1 \\
             \hline
            \multirow{1}{*}{$\text{Att}_\text{Co}$} & $5.80\pm5.19$ & 1.57 & $\times$43.12\\
            \hline
            \multirow{1}{*}{$\text{Att}_\text{Ny}$-4}  & $15.29\pm1.76$ & 8.41 & $\times$8.05\\
            \hline
            \multirow{1}{*}{$\text{Att}_\text{Ny}^\text{Fix}$-4}  &  $13.30\pm2.91$ & 1.24 & $\times$54.60\\
            \hline \hline
            \multirow{1}{*}{$\text{Att}_\text{CoNy}^\text{Cont}$-4 \textbf{(Ours)}} & $10.54\pm0.61$ & 0.47 & $\times$144,04 \\
            \hline
            \multirow{1}{*}{$\text{Att}_\text{CoNy}^\text{Fix}$-4 \textbf{(Ours)}} & $12.23\pm1.59$ & \textbf{0.23} & \textbf{$\times$294,35}\\
            \hline
        \end{tabular}
    \end{center}
\end{table}

\subsubsection{Electricity usage forecasting experiments}
We have also evaluated the performance of our model for electricity forecasting, using the Electricity Load Diagrams dataset \cite{Trindade15electricity}. It contains the electricity consumption of 370 customers over the span of 4 years, at intervals of 15 minutes. The given task is to predict the total electricity consumption in an hour time from the last given input. The input features contain the individual electricity consumptions, and temporal embeddings that track the day of the week and time of the day. All the features and the output are normalized to have zero mean and unit standard deviation per dimension, and we use an attention window of 64 tokens and one transformer block for all models.

Since the dataset is a single multi-variate time-series, we use the first 80\% of the series as the training set, the following 10\% as the validation set, and the last 10\% as the test set. We minimize the Huber loss function, and we use the AdamW optimizer \cite{Loshchilov19AdamW}.

The results of the experiments can be seen in Table~\ref{table:electricity}. All the models achieve a small normalized Mean Squared Error (MSE). The models that do not make use of the Nyström method achieve the lowest test loss. The continual models achieve a smaller loss than their non-continual counterparts. The overall computational overhead of the proposed Continual Nyströmformers is smaller than that of the baselines. We have also observed that for this task, the performance is slightly higher when using fixed landmarks.

\section{Conclusions} \label{sc:co}
In this paper, we introduced a new formulation of the Scaled Dot-product Attention based on the Nyström approximation that is suitable for Continual Inference. To do this, we derived the model updates of the Nyström-based SDA in a continual manner, and we proposed two ways tailored to processing continual stream data for determining the landmarks needed in the SDA approximation. The resulting model has a linear computational and memory cost with respect to the number of input tokens, and achieves faster inference time compared to competing models, while requiring comparable or lower memory.

This work highlights a logical process to obtain a continual version of a model that can potentially be used as an inspiration for other future adaptations of existing deep learning methods. Moreover, the proposed Continual Nyströmformers are not only more efficient at the individual layer than the previous approaches, but more Transformer layers can be stacked with a lower computational overhead than the Continual Transformers \cite{Hedegaard22continual_inference}, as all the intermediate layers are Nyströmformer layers \cite{Xiong21nystromformer} for our model, while the Continual Transformers use regular Transformer layers \cite{Vaswani17transformer}.

Experiments on Audio Classification, Online Action Detection and Electricity consumption forecasting show that the proposed model leads to a reduction of up to two orders of magnitude in the number of operations. The use of landmarks may translate into a slightly lower performance at the expense of a slightly lower performance. However, the performance results are variable depending on the task, and the models can benefit from using the Continual Nyströmformer attention. Based on these results, we believe that the Continual Nyströmformers are an interesting option to consider when creating deep learning models for the processing of temporal information in an online manner.

In many cases, the performance with few landmarks was similar to the full attention with no landmarks. We believe that this is caused by the big amount of redundant data present in most token matrices, and the Nyström-based models are able to generate attention outputs that are very similar to the outputs of the regular attention methods. We leave for future work a study in depth of this phenomenon. Other future works in the field can include the development of further methods in the family of Continual Inference; in particular, works that study and propose solutions to the limitation in the number of stacked Continual Transformers are especially relevant for the general success of Continual Transformers.

\appendix
\section{Computational costs calculation}\label{Appendix:comp}
This section shows the exact theoretical computational costs of the different attention modules. In particular, we show the cost of performing a single update during inference time on each of the variants of the Transformers discussed in this work. The real cost of these may be different based on the specific implementation and low-level operations.

In the following sections, $n$ makes reference to the sequence length, $d$ makes reference to the number of features and $m$ to the number of landmarks for the corresponding Nyström approximations.

\subsection{Base Transformer}\label{Ap:Transformer_comp}
The computational cost is $2n^2d + n^2 + nd + n$:
\begin{itemize}
    \item $nd$ from normalizing the matrix $Q$
    \item $n^2d$ to perform the multiplication $\frac{QK^T}{\sqrt{d}}$
    \item $n^2 + n$ to perform the softmax activation
    \item $n^2d$ to multiply the resulting matrix $\text{s}\left(\frac{QK^T}{\sqrt{d}}\right)$ against $V$
\end{itemize}

\subsection{Nyströmformer}\label{Ap:Nystromformer_comp}
The computational cost is $4ndm + 2nd + n + nm^2 + 2nm + dm^2 + 24m^3 + 22m^2 + 2m$:
\begin{itemize}
    \item $2nd$ from computing the landmark matrices $\tilde{Q}$ and $\tilde{K}$
    \item $2ndm + 2nm + n + m$ from computing the matrices $\text{s}\left(\frac{Q\tilde{K}^T}{\sqrt{d}}\right)$ and $\text{s}\left(\frac{\tilde{Q}K^T}{\sqrt{d}}\right)$
    \item $dm^2 + m^2 + m$ from computing the matrix $\text{s}\left(\frac{\tilde{Q}\tilde{K}^T}{\sqrt{d}}\right)$
    \item $24m^3 + 21m^2$ from computing the Moore-Penrose pseudo-inverse using the iterative algorithm \cite{Razavi14pinv}
    \item $nm^2 + 2ndm$ from the multiplication of the 3 low-rank matrices
\end{itemize}

\subsection{Nyströmformer with fixed landmarks}\label{Ap:TransformerFixed_comp}
The computational cost is $4ndm + nm^2 + 2nm + n + m$:
\begin{itemize}
    \item $2ndm + 2nm + n + m$ from computing the matrices $\text{s}\left(\frac{Q\tilde{K}^T}{\sqrt{d}}\right)$ and $\text{s}\left(\frac{\tilde{Q}K^T}{\sqrt{d}}\right)$
    \item $nm^2 + 2ndm$ from the multiplication of the 3 low-rank matrices
\end{itemize}

\subsection{Continual Retroactive Transformers}\label{Ap:CoTransformer_comp}
The computational cost is $7nd + 4n - 2d - 2$:
\begin{itemize}
    \item $nd + n - 1$ from computing $\rho\left(Q_\text{mem}, k_\text{old}\right)$ and $\rho\left(Q_\text{mem}, k_\text{new}\right)$
    \item $nd + n$ from computing the matrix $\rho\left(q_\text{new}, K\right)$
    \item $n$ from computing the vector $\phi(A)_\text{new}$
    \item $nd$ from computing the vector $AV_\text{new}$
    \item $n-1$ from updating $d_\text{mem}$
    \item $3nd - 2d$ from updating $AV_\text{mem}$
    \item $nd$ from computing $\phi(A)^{-1} \odot AV$
\end{itemize}

\subsection{Continual Single Output Transformer}\label{Ap:CoSiTransformer_comp}
The computational cost is $3nd + 2n$:
\begin{itemize}
    \item $nd + n$ from computing the matrix $a=\rho\left(q_\text{new}, K\right)$
    \item $nd$ from computing the matrix $aV$
    \item $n$ from computing the vector $\phi(a)$
    \item $nd$ from computing $aV\phi(a)^{-1}$
\end{itemize}

\subsection{Continual Retroactive Nyströmformers with Fixed Landmarks}
The computational cost is $ndm +6dm + m^2 + 6m$:
\begin{itemize}
    \item $dm + 3m$ from computing $(\beta_\phi)_\text{new}$
    \item $m^2$ from computing $(\beta_\phi)_\text{new}(\Gamma_\phi)^\dagger$
    \item $2dm + 2m$ from computing $\rho(\tilde{Q}, k_\text{old})$ and $\rho(\tilde{Q}, k_\text{new})$
    \item $m$ from updating $\phi(\Delta)^{-1}$
    \item $3dm$ from updating $\Delta V$
    \item $ndm$ to complete the attention computation
\end{itemize}

\subsection{Continual Single Output Nyströmformers with Fixed Landmarks}\label{Ap:CoSiFixedNystromformer_comp}
The computational cost is $7dm + m^2 + 6m$:
\begin{itemize}
    \item $dm + 3m$ from computing $(\beta_\phi)_\text{new}$
    \item $m^2$ from computing $(\beta_\phi)_\text{new}(\Gamma_\phi)^\dagger$
    \item $2dm + 2m$ from computing $\rho(\tilde{Q}, k_\text{old})$ and $\rho(\tilde{Q}, k_\text{new})$
    \item $m$ from updating $\phi(\Delta)^{-1}$
    \item $3dm$ from updating $\Delta V$
    \item $dm$ to complete the attention computation
\end{itemize}

\subsection{Continual Retroactive Nyströmformers with Continual Landmarks}
The computational cost is $ndm + 8nd + nm^2 + nm + 11n + 15dm + 2d + 24m^3 + 22m^2 + 22m$:
\begin{itemize}
    \item $dm + m$ from computing $\beta_\text{new}$
    \item $nd + n$ from computing $\rho(Q_\text{mem}, \tilde{k}_\text{new}^T)$
    \item $2d + m$ from updating $\Gamma$
    \item $nd + n + dm + m$ from updating $\Delta$
    \item $2nd + 4n + dm + 2m$ from updating $\phi(B)$
    \item $3dm + 6m$ from updating $\phi(\Gamma)$
    \item $nd + 2n + 2dm + 4m$ from updating $\phi(\Delta)$
    \item $24m^3 + 21m^2$ from computing the Moore-Penrose pseudo-inverse using the iterative algorithm \cite{Razavi14pinv}
    \item $nm^2 + nm + m^2$ from computing $B_\phi(\Gamma_\phi)^\dagger$
    \item $4dm + 2m$ from computing $\rho(\tilde{Q}, k_\text{old})v_\text{old}$ and $\rho(\tilde{Q}, k_\text{new})v_\text{new}$
    \item $nd + 2n + 2dm + 4m$ from computing $\phi(\Delta)$
    \item $2nd + n$ from computing $\Delta V$
    \item $dm + m$ from computing $\phi(\Delta)^{-1} \odot (\Delta V)$
    \item $ndm$ to complete the attention multiplication
\end{itemize}

\subsection{Continual Single Output Nyströmformers with Continual Landmarks}\label{Ap:CoSiNystromformer_comp}
The computational cost is $ndm + 3nd + n + 9dm + 2d + 24m^3 + 22m^2 + 13m$:
\begin{itemize}
    \item $24m^3 + 21m^2$ from computing the Moore-Penrose pseudo-inverse using the iterative algorithm \cite{Razavi14pinv}
    \item $2d + m$ from updating $(\Gamma_\phi)^\dagger$
    \item $2dm + m^2 + 3m$ from computing the updated $(\beta_\phi)_\text{new}(\Gamma_\phi)^\dagger$
    \item $3nd + n + 7dm + 9m$ from updating $\Delta V$
    \item $ndm$ to complete the attention multiplication
\end{itemize}

\section{Memory costs calculation}\label{Appendix:memory}
Computing the memory cost of the SDA module for Online Inference is not trivial. First, we need to distinguish the so-called valley cost, i.e., the minimum cost corresponding to the memory when no inference is performed and all the necessary matrices are stored, and the so-called peak cost, corresponding the maximum memory needed for performing calculations using intermediate matrices and storing the necessary matrices during an inference step. 

Since the main objective in Continual Inference is to improve processing speed by reducing redundant computations, optimizations (such as caching a compute result that can be used later resulting to a small memory increase in exchange for a computational cost reduction) are applied. Moreover, since specific implementation choices and low-level operations may lead to differences in the actual memory allocations of different implementations, in the following we consider the cases corresponding to the minimum possible cost.

\subsection{Base Transformer}\label{Ap:Transformer}
The valley cost is $3(nd-1)$:
\begin{itemize}
    \item $3(nd-1)$ from storing the matrices $Q_\text{mem}$, $K_\text{mem}$ and $V_\text{mem}$.
\end{itemize}

The peak cost is $n^2 + 4nd + 1$:
\begin{itemize}
    \item $3nd$ from storing the matrices $Q,K,V$
    \item $n^2$ from getting the matrix $\frac{QK^T}{\sqrt{d}}$
    \item $1$ from performing the softmax operation
    \item $nd$ from storing the output matrix
\end{itemize}

\subsection{Nyströmformer}\label{Ap:Nystromformer}
The valley cost is $3(nd-1)$:
\begin{itemize}
    \item $3(nd-1)$ from storing the matrices $Q_\text{mem}$, $K_\text{mem}$ and $V_\text{mem}$.
\end{itemize}

The peak cost is $4nd + 2nm + 2dm + 1 + 6m^2 + m$:
\begin{itemize}
    \item $3nd$ from storing the matrices $Q,K,V$
    \item $2dm$ from storing the matrices $\tilde{Q}, \tilde{K}$
    \item $nm + 1$ from storing the matrix $\text{s}(\frac{Q\tilde{K}^T}{\sqrt{d}})$
    \item $m^2$ from storing the matrix $\text{s}(\frac{\tilde{Q}\tilde{K}^T}{\sqrt{d}})$ (we reuse the +1 from the previous operation for the softmax).
    \item $5m^2 + m$ from computing the Moore-Penrose pseudo-inverse intermediate results
    \item $nm$ from storing the matrix $\text{s}(\frac{\tilde{Q}K^T}{\sqrt{d}})$
    \item $nd$ to compute the intermediate and final results.
\end{itemize}

\subsection{Nyströmformer with fixed landmarks}\label{Ap:TransformerFixed}
The valley cost is $3(nd-1) + 2dm + m^2$:
\begin{itemize}
    \item $3(nd-1)$ from storing the matrices $Q_\text{mem}$, $K_\text{mem}$ and $V_\text{mem}$
    \item $2dm$ from storing the matrices $\tilde{Q}, \tilde{K}$
    \item $m^2$ from storing the $(\Gamma_\phi)^\dagger$ matrix
\end{itemize}

The peak cost is $4nd + 2nm + 2dm + 2m^2 + 1$:
\begin{itemize}
    \item $3nd$ from storing the matrices $Q,K,V$
    \item $2dm$ from storing the matrices $\tilde{Q}, \tilde{K}$
    \item $m^2$ from storing the $(\Gamma_\phi)^\dagger$ matrix
    \item $nm + 1$ from storing the matrix $\text{s}(\frac{Q\tilde{K}^T}{\sqrt{d}})$
    \item $nm$ from storing the matrix $\text{s}(\frac{\tilde{Q}K^T}{\sqrt{d}})$
    \item $nd$ to compute the intermediate and final results.
\end{itemize}

\subsection{Continual Retroactive Transformers}\label{Ap:CoTransformer}
The valley cost is $4nd + n - d - 4$:
\begin{itemize}
    \item $3(nd-1)$ from storing the matrices $Q_\text{mem}$, $K_\text{mem}$ and $V_\text{mem}$
    \item $n-1$ from storing the vector $\phi(\hat{A}_\text{prev})$
    \item $nd - d$ from storing the matrix $\hat{AV}_\text{prev}$.
\end{itemize}

The peak cost is $5nd + 2n$:
\begin{itemize}
    \item $3nd$ from storing the matrices $Q$, $K$ and $V$
    \item $n$ from storing the vector $\phi(A)$
    \item $nd$ from storing the matrix $AV$.
    \item $n$ from storing the result of $\rho(Q_\text{mem},k_\text{old})$ (we can also use this to store $\rho(Q_\text{mem},k_\text{new})$ and $\rho(q_\text{new},K)$)
    \item $nd$ to store the final result.
\end{itemize}

\subsection{Continual Single Output Transformer}\label{Ap:CoSiTransformer}
The valley cost is $2(nd-1)$:
\begin{itemize}
    \item $2(nd-1)$ from storing the matrices $K_\text{mem}$ and $V_\text{mem}$
\end{itemize}

The peak cost is $2nd + n + 2d$:
\begin{itemize}
    \item $2nd$ from storing the matrices $K$ and $V$
    \item $d$ for storing $q_\text{new}$
    \item $n$ for storing $\rho(q_\text{new}, K)$
    \item $d$ to store the final result
\end{itemize}

\subsection{Continual Retroactive Nyströmformers with Fixed Landmarks}
The valley cost is $nm + 3dm + m^2 + m$:
\begin{itemize}
    \item $2dm$ from storing the matrices $\tilde{Q}$ and $\tilde{K}$
    \item $nm - m$ from storing the matrix $(B_\phi(\Gamma_\phi)^\dagger)_\text{mem}$
    \item $m^2$ from storing the matrix $(\Gamma_\phi)^\dagger$.
    \item $m$ from storing the cached vector $\phi(\Delta)^{-1}$
    \item $dm$ from storing the matrix $\Delta V$
\end{itemize}

The peak cost is $nd + nm + 3dm + m^2 + 2m$:
\begin{itemize}
    \item $2dm$ from storing the matrices $\tilde{Q}$ and $\tilde{K}$
    \item $nm$ from storing the matrix $(B_\phi(\Gamma_\phi)^\dagger)$
    \item $m^2$ from storing the matrix $(\Gamma_\phi)^\dagger$
    \item $m$ from storing the cached vector $\phi(\Delta)^{-1}$
    \item $dm$ from storing the matrix $\Delta V$
    \item $m$ to store $\rho(\tilde{Q}, k_\text{old})$
    \item $nd$ to store the final result
\end{itemize}

\subsection{Continual Single Output Nyströmformers with Fixed Landmarks}\label{Ap:CoSiFixedNystromformer}
The valley cost is $3dm + m^2 + m$:
\begin{itemize}
    \item $2dm$ from storing the matrices $\tilde{Q}$ and $\tilde{K}$
    \item $m^2$ from storing the matrix $(\Gamma_\phi)^\dagger$
    \item $m$ from storing the cached vector $\phi(\Delta)^{-1}$
    \item $dm$ from storing the matrix $\Delta V$
\end{itemize}

The peak cost is $3dm + d + m^2 + 2m$:
\begin{itemize}
    \item $2dm$ from storing the matrices $\tilde{Q}$ and $\tilde{K}$
    \item $m^2$ from storing the matrix $(\Gamma_\phi)^\dagger$
    \item $m$ from storing the cached vector $\phi(\Delta)^{-1}$
    \item $dm$ from storing the matrix $\Delta V$
    \item $m$ to store $\phi(\tilde{Q}, k_\text{old})$ and $(\beta_\phi(\Gamma_\phi)^\dagger)_\text{new}$
    \item $d$ to store the final result
\end{itemize}

\subsection{Continual Retroactive Nyströmformers with Continual Landmarks}
The valley cost is $3nd + 2nm + 4dm + 2m^2 - 2m - 6$:
\begin{itemize}
    \item $3(nd-1)$ from storing the matrices $Q_\text{mem}$, $K_\text{mem}$ and $V_\text{mem}$
    \item $2dm$ from storing the matrices $\tilde{Q}$ and $\tilde{K}$
    \item $nm - m$ from storing the matrix $(B_\phi(\Gamma_\phi)^\dagger)_\text{mem}$
    \item $m^2$ from storing the matrix $(\Gamma_\phi)^\dagger$
    \item $m$ from storing the cached vector $\phi(\Delta)^{-1}$
    \item $dm$ from storing the matrix $\Delta V$
    \item $n-1 + 2(m-1)$ for storing $\phi(\hat{B})_\text{prev}$, $\phi(\hat{\Gamma})_\text{prev}$ and $\phi(\hat{\Delta})_\text{prev}$
    \item $nm - n - m$ from storing $B_\text{prev}$
    \item $m^2 - 2m$ from storing $\Gamma_\text{prev}$
    \item $dm - m$ from storing $(\Delta V)_\text{prev}$
\end{itemize}

The peak cost is $4nd + 3nm + 2n + 4dm + 7m^2 + 4m$:
\begin{itemize}
    \item $3nd$ from storing the matrices $Q$, $K$ and $V$
    \item $2dm$ from storing the matrices $\tilde{Q}$ and $\tilde{K}$
    \item $nm$ from storing the matrix $B_\phi(\Gamma_\phi)^\dagger$
    \item $m^2$ from storing the matrix $(\Gamma_\phi)^\dagger$
    \item $m$ from storing the cached vector $\phi(\Delta)^{-1}$
    \item $dm$ from storing the matrix $\Delta V$
    \item $n + 2m$ for storing $\phi(\hat{B})$, $\phi(\hat{\Gamma})$ and $\phi(\hat{\Delta})$
    \item $nm$ from storing $B$
    \item $m^2$ from storing $\Gamma$
    \item $dm$ from storing $(\Delta V)$
    \item $n + m$ from 2 auxiliary vectors during computation
    \item $5m^2$ from the Moore-Penrose pseudo-inverse
    \item $nm$ to store the intermediate result $\Delta_\phi V$
    \item $nd$ to store the final computation
\end{itemize}

\subsection{Continual Single Output Nyströmformers with Continual Landmarks}\label{Ap:CoSiNystromformer}
The valley cost is $2nd + 4dm + 2m^2 - 2$:
\begin{itemize}
    \item $2(nd-1)$ from storing the matrices $K_\text{mem}$ and $V_\text{mem}$
    \item $2dm$ from storing the matrices $\tilde{Q}$ and $\tilde{K}$
    \item $m^2$ from storing the matrix $(\Gamma_\phi)^\dagger$
    \item $m$ from storing the cached vector $\phi(\Delta)^{-1}$
    \item $dm$ from storing the matrix $\Delta V$
    \item $2(m-1)$ for storing $\phi(\hat{\Gamma})_\text{prev}$ and $\phi(\hat{\Delta})_\text{prev}$
    \item $m^2 - 2m$ from storing $\Gamma_\text{prev}$
    \item $dm - m$ from storing $(\Delta V)_\text{prev}$
\end{itemize}

The peak cost is $2nd + nm + n + 4dm + d + 7m^2 + 4m$:
\begin{itemize}
    \item $2nd$ from storing the matrices $K$ and $V$
    \item $d$ from storing the vector $q_\text{new}$
    \item $2dm$ from storing the matrices $\tilde{Q}$ and $\tilde{K}$
    \item $m^2$ from storing the matrix $(\Gamma_\phi)^\dagger$
    \item $m$ from storing the cached vector $\phi(\Delta)^{-1}$
    \item $dm$ from storing the matrix $\Delta V$
    \item $2m$ for storing $\phi(\hat{\Gamma})$ and $\phi(\hat{\Delta})$
    \item $m^2$ from storing $\Gamma$
    \item $dm$ from storing $(\Delta V)$
    \item $n + m$ from 2 auxiliary vectors during computation
    \item $5m^2$ from the Moore-Penrose pseudo-inverse
    \item $nm$ to store the intermediate result $\Delta_\phi V$
\end{itemize}


\bibliographystyle{unsrt} 
\bibliography{refs}

\begin{thebibliography}{10}

\bibitem{Vaswani17transformer}
Ashish Vaswani, Noam Shazeer, Niki Parmar, Jakob Uszkoreit, Llion Jones, Aidan~N. Gomez, Lukasz Kaiser, and Illia Polosukhin.
\newblock Attention is all you need.
\newblock In {\em Advances in Neural Information Processing Systems}, pages 5998--6008, 2017.

\bibitem{Patwardhan23nlp}
Narendra Patwardhan, Stefano Marrone, and Carlo Sansone.
\newblock Transformers in the real world: {A} survey on {NLP} applications.
\newblock {\em Information}, 14(4):242, 2023.

\bibitem{Choi23genome}
Sanghyuk~Roy Choi and Minhyeok Lee.
\newblock Transformer architecture and attention mechanisms in genome data analysis: A comprehensive review.
\newblock {\em Biology}, 12(7):1033, 2023.

\bibitem{Ulhaq22ar_survey}
Anwaar Ulhaq, Naveed Akhtar, Ganna Pogrebna, and Ajmal Mian.
\newblock Vision transformers for action recognition: {A} survey.
\newblock {\em arXiv:2209.05700}, 2022.

\bibitem{Dosovitskiy21ViT}
Alexey Dosovitskiy, Lucas Beyer, Alexander Kolesnikov, Dirk Weissenborn, Xiaohua Zhai, Thomas Unterthiner, Mostafa Dehghani, Matthias Minderer, Georg Heigold, Sylvain Gelly, Jakob Uszkoreit, and Neil Houlsby.
\newblock An image is worth 16x16 words: Transformers for image recognition at scale.
\newblock In {\em International Conference on Learning Representations}, 2021.

\bibitem{Aggarwal22robotics}
Kriti Aggarwal, Sunil~K. Singh, Muskaan Chopra, Sudhakar Kumar, and Francesco Colace.
\newblock {\em Deep Learning in Robotics for Strengthening Industry 4.0.: Opportunities, Challenges and Future Directions}, volume 1030, chapter~1, pages 1--19.
\newblock Springer International Publishing, 2022.

\bibitem{Brohan23robotics}
Anthony Brohan, Noah Brown, Justice Carbajal, Yevgen Chebotar, Joseph Dabis, Chelsea Finn, Keerthana Gopalakrishnan, Karol Hausman, Alexander Herzog, Jasmine Hsu, Julian Ibarz, Brian Ichter, Alex Irpan, Tomas Jackson, Sally Jesmonth, Nikhil~J. Joshi, Ryan Julian, Dmitry Kalashnikov, Yuheng Kuang, Isabel Leal, Kuang{-}Huei Lee, Sergey Levine, Yao Lu, Utsav Malla, Deeksha Manjunath, Igor Mordatch, Ofir Nachum, Carolina Parada, Jodilyn Peralta, Emily Perez, Karl Pertsch, Jornell Quiambao, Kanishka Rao, Michael~S. Ryoo, Grecia Salazar, Pannag~R. Sanketi, Kevin Sayed, Jaspiar Singh, Sumedh Sontakke, Austin Stone, Clayton Tan, Huong~T. Tran, Vincent Vanhoucke, Steve Vega, Quan Vuong, Fei Xia, Ted Xiao, Peng Xu, Sichun Xu, Tianhe Yu, and Brianna Zitkovich.
\newblock {RT-1:} robotics transformer for real-world control at scale.
\newblock In {\em Robotics: Science and Systems XIX, Daegu, Republic of Korea, July 10-14, 2023}, 2023.

\bibitem{Miller24forecasting}
John~A. Miller, Mohammed Aldosari, Farah Saeed, Nasid~Habib Barna, Subas Rana, Ismailcem~Budak Arpinar, and Ninghao Liu.
\newblock A survey of deep learning and foundation models for time series forecasting.
\newblock {\em arXiv:2401.13912}, 2024.

\bibitem{bakhtiarnia2022single}
Arian Bakhtiarnia, Qi~Zhang, and Alexandros Iosifidis.
\newblock {Single-layer Vision Transformers for more accurate early exits with less overhead}.
\newblock {\em Neural Networks}, 153:461--473, 2022.

\bibitem{bakhtiarnia2024efficient}
Arian Bakhtiarnia, Qi~Zhang, and Alexandros Iosifidis.
\newblock {Efficient High-Resolution Deep Learning: A Survey}.
\newblock {\em {ACM} Computing Surveys}, 56(7):181:1--181:35, 2024.

\bibitem{Fan20transformer_depth}
Angela Fan, Edouard Grave, and Armand Joulin.
\newblock Reducing transformer depth on demand with structured dropout.
\newblock In {\em International Conference on Learning Representations}, 2020.

\bibitem{Xu24PCAFormer}
Jingjing Xu, Caesar Wu, Yuan{-}Fang Li, and Pascal Bouvry.
\newblock Transformer multivariate forecasting: Less is more?
\newblock {\em arXiv:2401.00230}, 2024.

\bibitem{Zhou21Informer}
Haoyi Zhou, Shanghang Zhang, Jieqi Peng, Shuai Zhang, Jianxin Li, Hui Xiong, and Wancai Zhang.
\newblock Informer: Beyond efficient transformer for long sequence time-series forecasting.
\newblock In {\em {AAAI} Conference on Artificial Intelligence}, pages 11106--11115, 2021.

\bibitem{Child19sparse_transformers}
Rewon Child, Scott Gray, Alec Radford, and Ilya Sutskever.
\newblock Generating long sequences with sparse transformers.
\newblock {\em arXiv:1904.10509}, 2019.

\bibitem{Beltagy20longformer}
Iz~Beltagy, Matthew~E. Peters, and Arman Cohan.
\newblock Longformer: The long-document transformer.
\newblock {\em arXiv:2004.05150}, 2020.

\bibitem{Choromanski21performer}
Krzysztof~Marcin Choromanski, Valerii Likhosherstov, David Dohan, Xingyou Song, Andreea Gane, Tam{\'{a}}s Sarl{\'{o}}s, Peter Hawkins, Jared~Quincy Davis, Afroz Mohiuddin, Lukasz Kaiser, David~Benjamin Belanger, Lucy~J. Colwell, and Adrian Weller.
\newblock Rethinking attention with performers.
\newblock In {\em International Conference on Learning Representations}, 2021.

\bibitem{Drineas05nystrom}
Petros Drineas and Michael~W. Mahoney.
\newblock On the nystr{\"{o}}m method for approximating a gram matrix for improved kernel-based learning.
\newblock {\em Journal on Machine Learning Research}, 6:2153--2175, 2005.

\bibitem{Yu17low_rank}
Xiyu Yu, Tongliang Liu, Xinchao Wang, and Dacheng Tao.
\newblock On compressing deep models by low rank and sparse decomposition.
\newblock In {\em {IEEE} Conference on Computer Vision and Pattern Recognition}, pages 67--76, 2017.

\bibitem{Xiong21nystromformer}
Yunyang Xiong, Zhanpeng Zeng, Rudrasis Chakraborty, Mingxing Tan, Glenn Fung, Yin Li, and Vikas Singh.
\newblock Nystr{\"{o}}mformer: {A} nystr{\"{o}}m-based algorithm for approximating self-attention.
\newblock In {\em AAAI Conference on Artificial Intelligence}, pages 14138--14148, 2021.

\bibitem{Yuan21ViT_training}
Li~Yuan, Yunpeng Chen, Tao Wang, Weihao Yu, Yujun Shi, Zihang Jiang, Francis E.~H. Tay, Jiashi Feng, and Shuicheng Yan.
\newblock Tokens-to-token vit: Training vision transformers from scratch on imagenet.
\newblock In {\em {IEEE/CVF} International Conference on Computer Vision}, pages 538--547, 2021.

\bibitem{Wu21CvT}
Haiping Wu, Bin Xiao, Noel Codella, Mengchen Liu, Xiyang Dai, Lu~Yuan, and Lei Zhang.
\newblock Cvt: Introducing convolutions to vision transformers.
\newblock In {\em {IEEE/CVF} International Conference on Computer Vision}, pages 22--31, 2021.

\bibitem{Khan22vision_transformers}
Salman~H. Khan, Muzammal Naseer, Munawar Hayat, Syed~Waqas Zamir, Fahad~Shahbaz Khan, and Mubarak Shah.
\newblock Transformers in vision: {A} survey.
\newblock {\em {ACM} Computing Surveys}, 54(10s):200:1--200:41, 2022.

\bibitem{Bertasius21space_time_attention}
Gedas Bertasius, Heng Wang, and Lorenzo Torresani.
\newblock Is space-time attention all you need for video understanding?
\newblock In {\em International Conference on Machine Learning}, pages 813--824, 2021.

\bibitem{Liu22video_swin}
Ze~Liu, Jia Ning, Yue Cao, Yixuan Wei, Zheng Zhang, Stephen Lin, and Han Hu.
\newblock Video swin transformer.
\newblock In {\em {IEEE/CVF} Conference on Computer Vision and Pattern Recognition}, pages 3192--3201, 2022.

\bibitem{Arnab21ViViT}
Anurag Arnab, Mostafa Dehghani, Georg Heigold, Chen Sun, Mario Lucic, and Cordelia Schmid.
\newblock Vivit: {A} video vision transformer.
\newblock In {\em {IEEE/CVF} International Conference on Computer Vision}, pages 6816--6826, 2021.

\bibitem{Fan21multiscaleViT}
Haoqi Fan, Bo~Xiong, Karttikeya Mangalam, Yanghao Li, Zhicheng Yan, Jitendra Malik, and Christoph Feichtenhofer.
\newblock Multiscale vision transformers.
\newblock In {\em {IEEE/CVF} International Conference on Computer Vision}, pages 6804--6815, 2021.

\bibitem{Hedegaard22continual_inference}
Lukas Hedegaard and Alexandros Iosifidis.
\newblock Continual inference: {A} library for efficient online inference with deep neural networks in pytorch.
\newblock In {\em European Conference on Computer Vision Workshops}, pages 21--34, 2022.

\bibitem{Schmidt19RNNs}
Robin~M. Schmidt.
\newblock Recurrent neural networks (rnns): {A} gentle introduction and overview.
\newblock {\em abs:1912.05911}, abs/1912.05911, 2019.

\bibitem{Shertinsky18LSTM}
Alex Sherstinsky.
\newblock Fundamentals of recurrent neural network {(RNN)} and long short-term memory {(LSTM)} network.
\newblock {\em abs:1808.03314}, abs/1808.03314, 2018.

\bibitem{Gu23Mambas}
Albert Gu and Tri Dao.
\newblock Mamba: Linear-time sequence modeling with selective state spaces.
\newblock {\em abs:2312.00752}, abs/2312.00752, 2023.

\bibitem{Hedegaard23cont_trans}
Lukas Hedegaard, Arian Bakhtiarnia, and Alexandros Iosifidis.
\newblock Continual transformers: Redundancy-free attention for online inference.
\newblock In {\em International Conference on Learning Representations}, 2023.

\bibitem{Hedegaard23GCNN}
Lukas Hedegaard, Negar Heidari, and Alexandros Iosifidis.
\newblock Continual spatio-temporal graph convolutional networks.
\newblock {\em Pattern Recognition}, 140(3):109528, 2023.

\bibitem{Hedegaard22Co3DCNN}
Lukas Hedegaard and Alexandros Iosifidis.
\newblock Continual 3d convolutional neural networks for real-time processing of videos.
\newblock In {\em European Conference Computer Vision}, pages 369--385, 2022.

\bibitem{Jeevan21xformers}
Pranav Jeevan and Amit Sethi.
\newblock Vision xformers: Efficient attention for image classification.
\newblock {\em arXiv:2107.02239}, 2021.

\bibitem{Scholkopf01kernels}
Bernhard Schölkopf and Alexander~J. Smola.
\newblock {\em Learning with Kernels: Support Vector Machines, Regularization, Optimization, and Beyond}.
\newblock The MIT Press, 2001.

\bibitem{Wang13nystrom}
Shusen Wang and Zhihua Zhang.
\newblock Improving {CUR} matrix decomposition and the nystr{\"{o}}m approximation via adaptive sampling.
\newblock {\em Journal of Machine Learning Research}, 14(1):2729--2769, 2013.

\bibitem{Patrick21Orthoformer}
Mandela Patrick, Dylan Campbell, Yuki~M. Asano, Ishan Misra, Florian Metze, Christoph Feichtenhofer, Andrea Vedaldi, and Jo{\~{a}}o~F. Henriques.
\newblock Keeping your eye on the ball: Trajectory attention in video transformers.
\newblock In {\em Advances in Neural Information Processing Systems}, pages 12493--12506, 2021.

\bibitem{Dutson23eventful_transformers}
Matthew Dutson, Yin Li, and Mohit Gupta.
\newblock Eventful transformers: Leveraging temporal redundancy in vision transformers.
\newblock In {\em {IEEE/CVF} International Conference on Computer Vision}, pages 16865--16877, 2023.

\bibitem{Lu21softmax_free}
Jiachen Lu, Jinghan Yao, Junge Zhang, Xiatian Zhu, Hang Xu, Weiguo Gao, Chunjing Xu, Tao Xiang, and Li~Zhang.
\newblock {SOFT:} softmax-free transformer with linear complexity.
\newblock In {\em Advances in Neural Information Processing Systems}, 2021.

\bibitem{Zhang23multi_resolution}
Yao Zhang, Yunpu Ma, Thomas Seidl, and Volker Tresp.
\newblock Adaptive multi-resolution attention with linear complexity.
\newblock In {\em International Joint Conference on Neural Networks, {IJCNN}}, pages 1--8. {IEEE}, 2023.

\bibitem{zhang2015scaling}
Kai Zhang, Liang Lan, James~T. Kwok, Slobodan Vucetic, and Bahram Parvin.
\newblock {Scaling up graph-based semisupervised learning via Prototype Vector Machines}.
\newblock {\em IEEE Transactions on Neural Networks and Learning Systems}, 26(3):444--457, 2015.

\bibitem{iosifidis2017class}
Alexandros Iosifidis and Moncef Gabbouj.
\newblock {Scaling Up Class-Specific Kernel Discriminant Analysis for Large-Scale Face Verification}.
\newblock {\em IEEE Transactions on Information Forensics and Security}, 11(11):2453--2465, 2016.

\bibitem{Zhang10clustered_nystrom}
Kai Zhang and James~T. Kwok.
\newblock Clustered nystr{\"{o}}m method for large scale manifold learning and dimension reduction.
\newblock {\em {IEEE} Trans. Neural Networks}, 21, 2010.

\bibitem{Razavi14pinv}
M.~Razavi, A.~Kerayechian, Mortaza Gachpazan, and Stanford Shateyi.
\newblock A new iterative method for finding approximate inverses of complex matrices.
\newblock {\em Abstract and Applied Analysis}, 2014(1):1--7, 2014.

\bibitem{Paszke19Pytorch}
Adam Paszke, Sam Gross, Francisco Massa, Adam Lerer, James Bradbury, Gregory Chanan, Trevor Killeen, Zeming Lin, Natalia Gimelshein, Luca Antiga, Alban Desmaison, Andreas K{\"{o}}pf, Edward~Z. Yang, Zachary DeVito, Martin Raison, Alykhan Tejani, Sasank Chilamkurthy, Benoit Steiner, Lu~Fang, Junjie Bai, and Soumith Chintala.
\newblock Pytorch: An imperative style, high-performance deep learning library.
\newblock In {\em Advances in Neural Information Processing Systems}, pages 8024--8035, 2019.

\bibitem{Goodfellow16DL}
Ian Goodfellow, Yoshua Bengio, and Aaron Courville.
\newblock {\em Deep Learning}.
\newblock MIT Press, 2016.

\bibitem{Stevens37mel}
Stanley~Smith Stevens, John Volkmann, and Edwin~Broomell Newman.
\newblock A scale for the measurement of the psychological magnitude pitch.
\newblock {\em The Journal of the Acoustical Society of America}, 8(3):185--190, 1937.

\bibitem{Choi16melhuman}
Keunwoo Choi, Gy{\"{o}}rgy Fazekas, and Mark~B. Sandler.
\newblock Automatic tagging using deep convolutional neural networks.
\newblock In {\em International Society for Music Information Retrieval Conference}, pages 805--811, 2016.

\bibitem{Palanisamy20audiocnn}
Kamalesh Palanisamy, Dipika Singhania, and Angela Yao.
\newblock Rethinking {CNN} models for audio classification.
\newblock {\em arXiv:2007.11154}, 2020.

\bibitem{Tzanetakis02GTZAN}
George Tzanetakis and Perry~R. Cook.
\newblock Musical genre classification of audio signals.
\newblock {\em {IEEE} Transactions on Speech and Audio Processing}, 10(5):293--302, 2002.

\bibitem{Simonyan15vgg}
Karen Simonyan and Andrew Zisserman.
\newblock Very deep convolutional networks for large-scale image recognition.
\newblock In {\em International Conference on Learning Representations}, 2015.

\bibitem{Hershey17AudioSet}
Shawn Hershey, Sourish Chaudhuri, Daniel P.~W. Ellis, Jort~F. Gemmeke, Aren Jansen, R.~Channing Moore, Manoj Plakal, Devin Platt, Rif~A. Saurous, Bryan Seybold, Malcolm Slaney, Ron~J. Weiss, and Kevin~W. Wilson.
\newblock {CNN} architectures for large-scale audio classification.
\newblock In {\em {IEEE} International Conference on Acoustics, Speech and Signal Processing}, pages 131--135, 2017.

\bibitem{DeGeest16OAD}
Roeland~De Geest, Efstratios Gavves, Amir Ghodrati, Zhenyang Li, Cees Snoek, and Tinne Tuytelaars.
\newblock Online action detection.
\newblock In {\em European Conference on Computer Vision}, pages 269--284, 2016.

\bibitem{Idrees17THUMOS}
Haroon Idrees, Amir~R. Zamir, Yu{-}Gang Jiang, Alex Gorban, Ivan Laptev, Rahul Sukthankar, and Mubarak Shah.
\newblock The {THUMOS} challenge on action recognition for videos "in the wild".
\newblock {\em Computer Vision and Image Understanding}, 155:1--23, 2017.

\bibitem{Lee23info_OAD}
Sumin Lee, Hyunjun Eun, Jinyoung Moon, Seokeon Choi, Yoonhyung Kim, Chanho Jung, and Changick Kim.
\newblock Learning to discriminate information for online action detection: Analysis and application.
\newblock {\em {IEEE} Transactions on Pattern Analysis and Machine Intelligence}, 45(5):5918--5934, 2023.

\bibitem{Wang19temporal_segment_networks}
Limin Wang, Yuanjun Xiong, Zhe Wang, Yu~Qiao, Dahua Lin, Xiaoou Tang, and Luc~Van Gool.
\newblock Temporal segment networks for action recognition in videos.
\newblock {\em {IEEE} Transactions on Pattern Analysis and Machine Intelligence}, 41(11):2740--2755, 2019.

\bibitem{Heilbron15ActivityNet}
Fabian~Caba Heilbron, Victor Escorcia, Bernard Ghanem, and Juan~Carlos Niebles.
\newblock Activitynet: {A} large-scale video benchmark for human activity understanding.
\newblock In {\em {IEEE} Conference on Computer Vision and Pattern Recognition}, pages 961--970, 2015.

\bibitem{Carreira17Kinetics}
Jo{\~{a}}o Carreira and Andrew Zisserman.
\newblock Quo vadis, action recognition? {A} new model and the kinetics dataset.
\newblock In {\em {IEEE} Conference on Computer Vision and Pattern Recognition}, pages 4724--4733, 2017.

\bibitem{Trindade15electricity}
Artur Trindade.
\newblock {ElectricityLoadDiagrams20112014}.
\newblock UCI Machine Learning Repository, 2015.
\newblock {DOI}: https://doi.org/10.24432/C58C86.

\bibitem{Loshchilov19AdamW}
Ilya Loshchilov and Frank Hutter.
\newblock Decoupled weight decay regularization.
\newblock In {\em 7th International Conference on Learning Representations, {ICLR}}, 2019.

\end{thebibliography}



\end{document}